%% file: main.tex
\documentclass[10pt,twocolumn,letterpaper]{article}

\usepackage[pagenumbers]{config/cvpr} %

\input{config/packages}

\input{config/commands}

\def\cvprPaperID{914} %
\def\confName{CVPR}
\def\confYear{2023}

\begin{document}

\title{\vspace{-0.5 em}\outTitleFULL\vspace{-0.5 em}}

\input{config/authors}

\twocolumn[{%
\renewcommand\twocolumn[1][]{#1}%
\maketitle
\input{TEX_figures/0_teaser}

}]

\subfile{sections_Paper/0_abstract}

\section{Introduction}
\label{sec:intro}

\subfile{sections_Paper/1_intro}

\section{Related Work}
\label{sec:related_work}
\subfile{sections_Paper/2_related}

\section{Method}
\label{sec:method}
\subfile{sections_Paper/3_method}

\section{Experiments}
\label{sec:experiments}
\subfile{sections_Paper/4_experiments}

\section{Conclusion}
\label{sec:conclusion}
\subfile{sections_Paper/5_conclusion}

\label{sec:acknowledgements}
\subfile{sections_Paper/6_acknowledgements}

\appendix
{\noindent\LARGE\textbf{Supplementary Material}}
\newline
\renewcommand{\thefigure}{S.\arabic{figure}}
\renewcommand{\thetable}{S.\arabic{table}}
\renewcommand{\theequation}{S.\arabic{equation}}
\setcounter{figure}{0}
\setcounter{table}{0}
\setcounter{equation}{0}

\input{supplementary/sections_Supmat/moyo_details}

\section{Method}

\input{supplementary/sections_Supmat/stability_loss_justification}
\input{supplementary/sections_Supmat/m_naive_m_trig}

\section{Experiments}

\input{supplementary/sections_Supmat/implementation_details}

\input{supplementary/sections_Supmat/bose_calculation}

\input{supplementary/sections_Supmat/extra_qualitative_results}

\section{Stability Evaluation via Physics Simulation}
\input{supplementary/sections_Supmat/physics_simulation}

\section{Evaluation of Biomechanical Elements}
\input{supplementary/sections_Supmat/biomechanical_eval}

\input{supplementary/sections_Supmat/04_smplifyx_IPMAN_O}

\input{supplementary/sections_Supmat/3dpw_eval}

\pagebreak

{\small
\bibliographystyle{config/ieee_fullname}
\bibliography{config/BIB}
\balance
}

\end{document}

%% file: config/packages.tex
\usepackage{graphicx}

\usepackage{amsmath}
\usepackage{amssymb}
\usepackage{epsfig}
\usepackage[table,usenames,dvipsnames]{xcolor}
\usepackage{datetime2}
\usepackage{bm}
\usepackage{color,soul}          %
\usepackage{microtype}      %
\usepackage{dcolumn}        %
\usepackage{url}            %
\usepackage{booktabs}       %
\usepackage{amsfonts}       %
\usepackage{nicefrac}       %
\usepackage{enumitem}
\usepackage{multirow}
\usepackage{float}
\usepackage{currfile}
\usepackage{algorithm}
\usepackage[noend]{algpseudocode}
\usepackage{amsthm}
\usepackage{subfiles}
\usepackage{adjustbox}
\usepackage{subcaption}
\usepackage{makecell}
\usepackage[normalem]{ulem}
\usepackage{caption}
\usepackage[flushmargin,para]{footmisc}
\usepackage{pifont}%
\usepackage{pifont}%
\usepackage{adjustbox}
\usepackage[pagebackref,breaklinks,colorlinks]{hyperref} %

\usepackage[accsupp]{axessibility}  %

\usepackage{wrapfig}

\usepackage{hhline}

\usepackage{balance}

\usepackage{fontawesome}

\usepackage[toc,page]{appendix}
\usepackage[accsupp]{axessibility}

\usepackage[capitalize]{cleveref}
\crefname{section}{Sec.}{Secs.}
\Crefname{section}{Section}{Sections}
\Crefname{table}{Table}{Tables}
\crefname{table}{Tab.}{Tabs.}

%% file: config/commands.tex
\usepackage[normalem]{ulem}
\xspaceaddexceptions{]\}}

\newcommand{\qheading}[1]{\noindent\textbf{#1.}}

\newcommand{\zheading}[1]{\textbf{#1.}}

\newcommand{\nameCOLOR}[1]{\textcolor{black}{#1}} %
\newcommand{\nameMethod}{\nameCOLOR{\mbox{IPMAN}}\xspace}
\newcommand{\IPMAN}{\nameMethod}
\newcommand{\nameMethodLONG}{\nameCOLOR{Intuitive-Physics-based huMAN}}
\newcommand{\nameMethodR}{\nameCOLOR{\mbox{IPMAN-R}}\xspace}
\newcommand{\nameMethodO}{\nameCOLOR{\mbox{IPMAN-O}}\xspace}
\newcommand{\nameMethodOnew}{\nameCOLOR{\mbox{IPMAN-O\NEWWW{*}}}\xspace}

\newcommand{\nameYOGIData}{\nameCOLOR{\mbox{MoYo}}\xspace}
\newcommand{\nameYOGIDataBig}{\nameCOLOR{\mbox{\mocap Yoga}}\xspace}
\newcommand{\yogiIOI}{\nameYOGIData}
\newcommand{\outTitleFULL}{\threeD Human Pose Estimation via Intuitive Physics}

\newcommand{\ip}{\nameCOLOR{IP}\xspace}
\newcommand{\IP}{\ip}
\newcommand{\IPterms}{\nameCOLOR{{\ip terms}}\xspace}

\newcommand{\NEWWW}[1]{\xspace{\textcolor{black}{#1}}\xspace}

\newcommand{\rich}{\mbox{\nameCOLOR{RICH}}\xspace}

\newcommand{\supmat}{\textcolor{black}{{Sup.~Mat.}}\xspace}

\newcommand{\highlightNUMB}[1]{\textcolor{black}{#1}} %
\newcommand{\highlightSYMBOLS}[1]{\textcolor{black}{#1}} %
\newcommand{\bs}[1]{\boldsymbol{{#1}}}

\newcommand{\TODO}[1]{\textcolor{black}{#1}}

\newcommand{\newtext}[1]{\textcolor{black}{#1}}
\newcommand{\newcamready}[1]{\textcolor{black}{#1}}
\newcommand{\optional}[1]{\textcolor{black}{#1}}
\newcommand{\deadline}[1]{\textcolor{black}{#1}}
\newcommand{\SUBSPRINT}[1]{\deadline{#1}}

\newcommand{\smpl}{\mbox{SMPL}\xspace}
\newcommand{\smplx}{\mbox{SMPL-X}\xspace}
\newcommand{\smplX}{\smplx}
\newcommand{\hmr}{\mbox{HMR}\xspace}

\newcommand{\hps}{\mbox{HPS}\xspace}
\newcommand{\HPS}{\hps}

\newcommand{\threedpw}{\mbox{3DPW}\xspace}

\newcommand{\humanThreeSixFull}{\mbox{Human3.6M}\xspace}
\newcommand{\humanthreesix}{\mbox{H3.6M}\xspace}

\newcommand{\mpjpe}{\mbox{\highlightSYMBOLS{MPJPE}}\xspace}
\newcommand{\pampjpe}{\mbox{\highlightSYMBOLS{PA-MPJPE}}\xspace}
\newcommand{\pve}{\mbox{\highlightSYMBOLS{PVE}}\xspace}

\newcommand{\physCOLOR}[1]{\textcolor{black}{#1}} %
\newcommand{\physExplainCOM}{{Center of Mass}\xspace}
\newcommand{\physExplainBOS}{{Base of Support}\xspace}
\newcommand{\physExplainCOP}{{Center of Pressure}\xspace}
\newcommand{\physBOS}{\physCOLOR{\mbox{BoS}}\xspace}
\newcommand{\physBOSE}{\physCOLOR{\mbox{BoSE}}\xspace}
\newcommand{\physCOP}{\physCOLOR{\mbox{CoP}}\xspace}
\newcommand{\physCOM}{\physCOLOR{\mbox{CoM}}\xspace}
\newcommand{\physCOMv}{\physCOLOR{\mbox{pCoM}}\xspace}

\newcommand{\physCOMp}{\physCOMv}

\newcommand{\rgb}{\mbox{RGB}\xspace}

\renewcommand{\etc}{etc\xspace}
\renewcommand{\etal}{\mbox{et al.}\xspace}
\renewcommand{\ie}{\mbox{i.e.}\xspace}
\renewcommand{\eg}{\mbox{e.g.}\xspace}
\renewcommand{\wrt}{\mbox{w.r.t.}\xspace}

\newcommand{\pose}{\boldsymbol{\theta}\xspace}
\newcommand{\shape}{\boldsymbol{\beta}\xspace}

\newcommand{\twoD}{{2D}\xspace}
\newcommand{\sixD}{{6D}\xspace}
\newcommand{\threeD}{\xspace{3D}\xspace}

\newcommand{\mocap}{\mbox{MoCap}\xspace}

\newcommand{\smplifyXMC}{\mbox{SMPLify-XMC}\xspace}
\newcommand{\smplifyx}{\mbox{SMPLify-X}\xspace}
\newcommand{\smplifyX}{\smplifyx}
\newcommand{\groundtruth}{{ground-truth}\xspace}

\newcommand{\SOTA}{\mbox{SOTA}\xspace}

\newcommand{\reffig}[1]{\cref{#1}}

\newcommand{\refFig}[1]{\Cref{#1}}
\newcommand{\reftab}[1]{\cref{#1}}
\newcommand{\refTab}[1]{\Cref{#1}}

\newcommand{\refsec}[1]{\cref{#1}}

\newcommand{\volume}{\mathcal{V}}
\newcommand{\meshSurfPointi}{v_i}
\newcommand{\meshPartSET}{\mathcal{P}}
\newcommand{\meshPart}{P}
\newcommand{\meshParti}{\meshPart_i}
\newcommand{\meshPartvi}{\meshPart_{\meshSurfPointi}}
\newcommand{\volumePart}{\volume^\meshPartSET}
\newcommand{\volumeParti}{\volume^{\meshParti}}
\newcommand{\volumePartvi}{\volume^{\meshPartvi}}

\newcommand{\camtransl}{\mathbf{t}^c}
\newcommand{\camrot}{\mathbf{R}^c}
\newcommand{\bodytransl}{\mathbf{t}^b}
\newcommand{\bodyori}{\mathbf{R}^b}

\definecolor{lightgray}{gray}{0.97}
\definecolor{lightblue}{rgb}{0.93,0.95,1.0}

\definecolor{GreenColor}{rgb}{0.137,0.573,0.565}
\definecolor{OrangeColor}{rgb}{0.914,0.541,0.0.141}
\definecolor{PurpleColor}{rgb}{0.5,0,0.7}

%% file: config/authors.tex
\author{
Shashank Tripathi$^1$ \quad Lea M\"{u}ller$^1$ \quad Chun-Hao P. Huang$^1$ \quad Omid Taheri$^1$ \\ Michael J. Black$^1$ \quad Dimitrios Tzionas$^2$\footnotemark[1] \\
{\small
$^1$Max Planck Institute for Intelligent Systems, T{\"u}bingen, Germany \quad
$^2$University of Amsterdam, the Netherlands
}\\
{\tt\small \{stripathi,  lmueller2, chuang2, otaheri, black\}@tue.mpg.de \quad {d.tzionas@uva.nl}}\\
}

%% file: TEX_figures/0_teaser.tex
\begin{center}
    \centering
    \vspace{-1.0 em}
    \includegraphics[width=\linewidth]{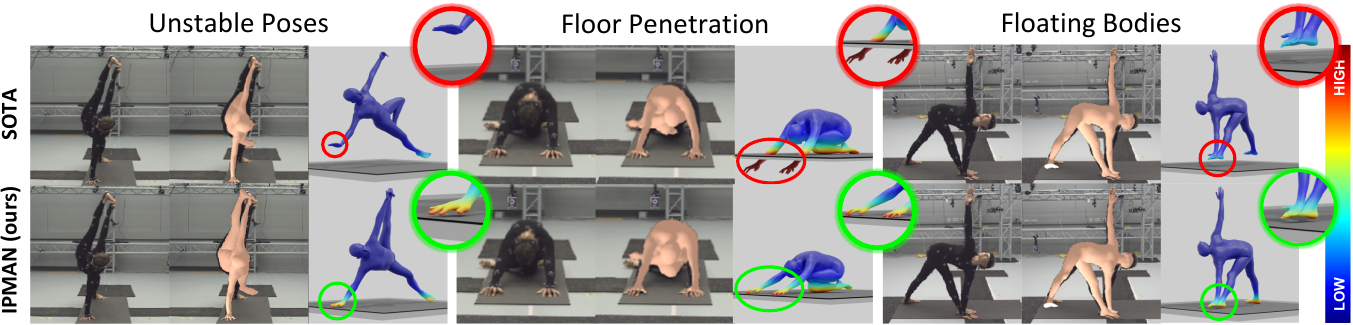}
    \vspace{-1.7 em}
    \captionsetup{type=figure}
    \caption{Estimating a \threeD body from an image is ill-posed.
        A recent, representative, optimization method~\cite{Mueller:CVPR:2021}        %
        produces bodies that are in 
        unstable poses,          %
        penetrate the floor,      %
        or hover above it.       %
        In contrast, \nameMethod %
        estimates a \threeD body 
        that is physically {\em plausible}.  
        To achieve this, \nameMethod uses novel {\em intuitive-physics} (\IP) terms that exploit inferred
        \emph{pressure} heatmaps on the body,
        the  \emph{\physExplainCOP} (\physCOP), and 
        the body's  \emph{\physExplainCOM} (\physCOM). %
        Body heatmap colors encode per-vertex pressure.
}
    \label{figure:teaser_new}
    \vspace{+1.3 em}
\end{center}%

%% file: sections_Paper/0_abstract.tex
\begin{abstract}
\vspace{-0.2em}
Estimating \threeD humans from images often produces implausible bodies that lean, float, or penetrate the floor.
Such methods ignore the fact that bodies are typically
supported by the scene. 
A physics engine 
can be used 
to enforce physical plausibility, 
but these are
not differentiable, rely on unrealistic proxy bodies, and are difficult to integrate into existing optimization and learning frameworks.
In contrast, we exploit
novel \TODO{intuitive-physics} (\IP) terms that can be inferred from a 3D \smpl body interacting with the scene.
Inspired by biomechanics, we infer
the \emph{pressure} heatmap on the body,
 the 
\emph{\physExplainCOP} (\physCOP) from the heatmap, and the \smpl body's 
\emph{\physExplainCOM} (\physCOM).
With these, we develop \nameMethod, to estimate a \threeD body 
from a color image in a ``stable'' configuration by encouraging %
plausible floor contact and  overlapping \physCOP and \physCOM. 
Our \IPterms are intuitive, easy to implement, fast to compute, differentiable, and can be integrated into 
existing
optimization and regression methods. %
We evaluate \nameMethod on standard datasets and
\nameYOGIData, a new dataset with synchronized multi-view images, ground-truth 3D bodies with complex poses, body-floor contact, \physCOM and pressure. 
\nameMethod produces
more plausible results than the state of the art, 
improving accuracy for static poses, %
while not hurting dynamic ones. %
Code and data are available for research at \url{https://ipman.is.tue.mpg.de}. %
\end{abstract}
\enlargethispage{2.0 em} %

\renewcommand{\thefootnote}{\fnsymbol{footnote}}
\footnotetext[1]{This work was mostly performed at MPI-IS.}

%% file: sections_Paper/1_intro.tex
To understand humans and their actions, computers need automatic methods to reconstruct the body in \threeD. %
Typically, the problem entails estimating the \threeD human pose and shape (\hps) from one or more color images. 
State-of-the-art (\SOTA) methods~\cite{Kocabas2021pare, ROMP:ICCV:2021, Zhang2021PyMAF3H, li2022cliff} have made rapid progress, estimating \threeD humans that \emph{align} well with image features in the camera view. 
Unfortunately, the camera view can be deceiving. 
When viewed from other directions, or when placed in a \threeD scene, the estimated 
bodies are often physically implausible:
they lean, hover, or penetrate the ground
(see \reffig{figure:teaser_new}~top). 
This is because most \SOTA methods reason about humans  \emph{in isolation}; %
they ignore that people move in a scene, interact with it, and receive physical support by contacting it. %
This is a \emph{deal-breaker} for inherently \threeD applications, such as biomechanics, augmented/virtual reality (AR/VR) and the ``metaverse''; 
these need humans to be reconstructed faithfully and \emph{physically plausibly} with respect to the scene.
For this, we need a method that estimates the \threeD human on a ground plane from a color image 
in a configuration that is \emph{physically ``stable''}.

This is naturally related to reasoning about physics and support.
There exist many physics simulators~\cite{havok,pyBullet,physX} for games, movies, or industrial simulations, and using these for plausible \HPS inference is increasingly popular \cite{Shimada2020PhysCapTOG, RempeContactDynamics2020, yuan2021simpoe}. 
However, existing simulators come with two significant problems: 
\highlightNUMB{(1)}
They are \newtext{typically} non-differentiable \emph{black boxes}, making them incompatible with existing optimization and learning frameworks. 
Consequently, most methods~\cite{peng2018deepmimic, yuan2021simpoe, yuan20183d} use them with reinforcement learning  %
to evaluate whether a certain input has the desired outcome, but with no ability to reason about how changing inputs affects the outputs. 
\highlightNUMB{(2)}
They rely on an unrealistic proxy body model for computational efficiency; 
bodies are represented as 
groups of rigid \threeD shape primitives. 
Such proxy models are crude approximations of human bodies, which, in reality, are much more complex and deform non-rigidly when they move and interact. 
Moreover, proxies need {\em a priori} known body dimensions that are kept fixed during simulation. 
Also, these proxies differ significantly from the %
\threeD body models \cite{Joo2018_adam,SMPL:2015,xu2020ghum} used by \SOTA \HPS methods. 
Thus, current physics simulators are too limited for use in \HPS.

\begin{figure}[t]
\centering
\vspace{-0.5 em}
\includegraphics[width=\linewidth]{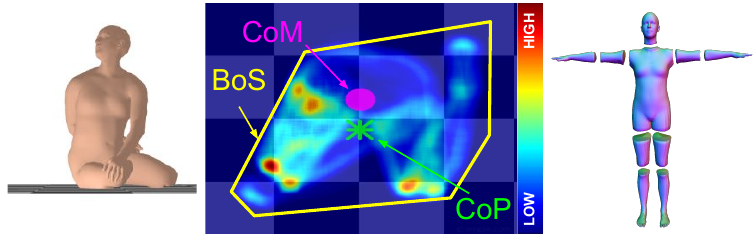}
\vspace{-2.0 em}
\caption{
    (1)
    A \smpl mesh sitting. %
    (2)
    The inferred \TODO{pressure map} on the ground (color-coded heatmap), %
    \physCOP (green), 
    \physCOM (pink), and 
    \physExplainBOS (\physBOS, yellow polygon). 
    (3)
    Segmentation of \smpl into $N_P = 10$ parts, used for computing \physCOM; see \refsec{subsec:elements_of_stability_analysis}. 
}
\vspace*{-1.0 em}
\label{figure:no_bos_loss}
\end{figure}

What we need, instead, is a solution that is fully differentiable, uses a realistic body model, and seamlessly integrates physical reasoning into \HPS methods (both optimization- and regression-based). %
To this end, instead of using full physics simulation, we introduce novel intuitive-physics (\IP) terms that are simple, differentiable, and compatible with a body model like \smpl~\cite{SMPL:2015}. 
Specifically, we define terms that exploit \newtext{an inferred}
\emph{pressure} heatmap of the body on the ground plane, the %
\emph{\physExplainCOP} (\physCOP) that arises from the heatmap, and the \smpl body's 
\emph{\physExplainCOM} (\physCOM) projected on the floor; 
see \reffig{figure:no_bos_loss} for a %
visualization. 
Intuitively, bodies whose \physCOM lie close to their \physCOP are more \emph{\TODO{stable}} than ones with a \physCOP that is further away (see \reffig{figure:ipman_qual_result_both}); the former suggests a \emph{static pose}, \eg standing or holding a yoga pose, while the latter %
a \emph{dynamic pose}, \eg, 
walking.

We use these intuitive-physics terms in two ways.
First, we incorporate them in an objective function that extends \smplifyXMC \cite{Mueller:CVPR:2021} to optimize for body poses that are stable.
We also incorporate 
the same 
terms in the training loss for an \HPS regressor, called \nameMethod (\nameMethodLONG).
In both formulations, the intuitive-physics terms encourage estimates of body shape and pose that have sufficient ground contact, while penalizing interpenetration and encouraging an overlap of the \physCOP and \physCOM.

Our intuitive-physics formulation is inspired by work in biomechanics  \cite{pai2003movement,hof2007equations,hof2008extrapolated}, which characterizes
the stability of 
humans 
in terms of relative positions between the 
\physCOP, the \physCOM, and the
\emph{\physExplainBOS} (\physBOS).
The \physBOS is defined as the convex hull of all contact regions on the floor (\reffig{figure:no_bos_loss}).
Following past work \cite{Shimada2020PhysCapTOG,scott2020image,barnett2011perceived}, we use the ``inverted pendulum'' model~\cite{winter1995abc,winter1995human} for body balance; this considers poses as stable if the gravity-projected \physCOM onto the floor lies inside the \physBOS. 
Similar ideas are explored by Scott et al.~\cite{scott2020image} but they focus on predicting a foot pressure heatmap from \twoD or \threeD %
body joints.
We go significantly further to exploit stability in training an \HPS regressor.
This requires two technical novelties.

The first involves computing \physCOM. 
To this end, we uniformly sample 
\TODO{points on \smpl's surface}, and calculate each body part's volume. 
Then, we compute \physCOM as the average of all uniformly \TODO{sampled points} weighted by the corresponding part volumes. 
We denote this as \physCOMv, standing for ``part-weighted \physCOM''. 
Importantly, \physCOMv takes into account \smpl's shape, pose, and all blend shapes, while it is also computationally efficient and differentiable.

The second involves estimating \physCOP directly from the image, %
without access to a pressure sensor. 
Our key insight is that the soft tissues of human bodies deform under pressure, \eg, the buttocks deform when sitting. 
However, \smpl does not model this deformation; %
it {\em penetrates} the ground instead of deforming.
We use the penetration depth as a proxy for
pressure~\cite{Rogez2015everyday};
deeper 
penetration means higher pressure.
\newtext{With this, we estimate a pressure field on \smpl's mesh and compute the \physCOP as the pressure-weighted average of the
\TODO{surface points}.} 
Again this is differentiable.

For evaluation, we use a standard \HPS benchmark %
(\humanThreeSixFull~\cite{ionescupapavaetal2014}), %
but also the \rich~\cite{RICH} dataset. 
However, these datasets have limited interactions with the floor. 
\newtext{We thus capture a novel dataset, \nameYOGIData, of challenging yoga poses, with synchronized multi-view video,
\groundtruth \smplX~\cite{Pavlakos2019_smplifyx} meshes,}
\newtext{pressure sensor measurements, and body \physCOM}.
\nameMethod, in both of its forms, and across all datasets,
produces more accurate and stable \threeD bodies than the state of the art. 
Importantly, we find that \nameMethod improves accuracy for static poses, while not hurting dynamic ones. This makes \nameMethod applicable to everyday motions. 

\newtext{To summarize: 
(1) We develop \nameMethod, the first \hps method that integrates intuitive physics. %
(2) We infer biomechanical properties such as \physCOM, \physCOP and body pressure. %
(3) We \TODO{define} novel %
\emph{intuitive-physics} terms that can be easily integrated into \hps methods. 
(4) We create \nameYOGIData, a dataset that uniquely has complex poses, multi-view video, and ground-truth bodies, pressure, and \physCOM.   %
(5) We show that our \IPterms 
improve \HPS accuracy and physical plausibility. %
(6)~Data and code are available for research.} %

%% file: sections_Paper/2_related.tex
\zheading{\threeD Human Pose and Shape (\hps) from images}
Existing methods 
fall 
into two major categories: 
(1) non-parametric methods that reconstruct a free-form body representation, 
\eg, joints \cite{Akhter:CVPR:2015,Mehta2017Monocular3H,mehta2017vnect} or 
vertices \cite{lin2021end-to-end,Moon_2020_ECCV_I2L-MeshNet,Zeng_2020_CVPR}, and
(2) parametric methods that use statistical body models \cite{Gler2019HoloPoseH3, anguelov2005scape,Joo2018_adam,SMPL:2015,Pavlakos2019_smplifyx,xu2020ghum, zanfir2020weakly}. 
The latter methods focus on various aspects, such as 
expressiveness \cite{Choutas2020_expose,feng2021pixie,Pavlakos2019_smplifyx,rong2021frankmocap,Xiang_2019_CVPR}, 
clothed bodies \cite{corona2021smplicit,xiang2020monoclothcap,xiu2023econ},
videos \cite{GuanXWNY21,VIBE:CVPR:2020,TripathiRTA20,zeng2022smoothnet}, and 
multi-person scenarios \cite{jiang2020multiperson,ROMP:ICCV:2021,zhang2021bmp}, to name a few.

Inference is done by
either optimization or regression.
Optimization-based methods \cite{bogo2016keep,fan2021revitalizing,Pavlakos2019_smplifyx,Xiang_2019_CVPR,xiang2020monoclothcap} fit a body model to image evidence, such as joints \cite{cao2019openpose}, dense vertex correspondences \cite{alp2018densepose} or \twoD segmentation masks \cite{gong2019graphonomy}.
Regression-based methods \cite{Kanazawa2018_hmr, li2022cliff, Kolotouros2019ConvolutionalMR, Yu2019HumanMR, Zhang2021PyMAF3H, Khirodkar2022OccludedHM, Zhou2021MonocularRF, Zhang2020ObjectOccludedHS} use a loss similar to the objective function of optimization methods to train a %
network to infer body model parameters.
Several methods combine optimization and regression in a training loop
 \cite{kolotouros2019spin, Mueller:CVPR:2021, li2021hybrik}.
Recent methods \cite{joo2021eft,GuanXWNY21}
fine-tune pre-trained networks at test time w.r.t.~an image or a sequence, retaining
flexibility (optimization) while being less sensitive to initialization (regression). %

Despite their success, these methods reason about the human in ``isolation'', without taking the surrounding scene into account; 
see \cite{tian2022hmrsurvey,zheng2022deep} for a comprehensive 
review.  %

\zheading{Contact-only scene constraints} 
A common way of %
using 
scene information is to consider body-scene \emph{contact} \cite{Hassan2019prox,Hassan2021posa,rempe2021humor,weng2020holistic,zanfir2018monocular,zhang2020phosa,Zhang_2021_ICCV,zou2020reducing, xie22chore, yi2022mime, fan2023arctic, chen2023hot}.
Yamamoto \etal~\cite{YamamotoCVPR2000} and others \cite{Hassan2019prox,zanfir2018monocular,zhang2020phosa,fieraru2021remips,rueegg2023bite} ensure that %
estimated bodies 
have plausible scene contact. %
For 
videos, encouraging foot-ground contact %
reduces foot skating %
\cite{Ikemoto2006,rempe2021humor,shi2020motionet,Zhang_2021_ICCV,zou2020reducing}.
Weng \etal~\cite{weng2020holistic} %
use contact in %
estimating the pose and scale of scene objects,
while 
Villegas \etal~\cite{Villegas_2021_ICCV} preserve 
self- and ground contact 
\newtext{for motion retargeting.} 

These methods typically take two steps: 
(1) detecting contact areas \newtext{on the body and/or scene} %
    and 
(2) minimizing the distance between \newtext{these}. %
Surfaces are typically assumed to be in contact if their %
distance is below a threshold and their relative motion is small
\cite{Hassan2019prox,RICH,zanfir2018monocular,zhang2020phosa}.

Many methods only consider contact between the ground and the foot joints
\cite{zou2020reducing, RempeContactDynamics2020} 
or \newtext{other} end-effectors \cite{rempe2021humor}.
In contrast, \IPMAN uses the 
\newtext{full \threeD body surface} 
and exploits this to compute the pressure, \physCOP and \physCOM.
Unlike \TODO{binary contact}, this is differentiable, making the \IPterms  useful for training 
\HPS 
regressors. 

\zheading{Physics-based scene constraints} 
Early work uses physics to estimate walking \cite{Brubaker:CVPR:2009,Brubaker:IJCV:2010} or full body motion \cite{Vondrak:CVPR:2008}.
Recent methods \cite{RempeContactDynamics2020,Shimada2020PhysCapTOG,Shimada2021PhysAwareTOG,yuan2021simpoe,Xie_2021_ICCV, gartner2022trajectory, gartner2022diffphy} 
regress \threeD humans and then refine them 
through \emph{physics-based optimization}.
Physics is 
\newtext{used} 
for two primary reasons: 
(1)
to regularise dynamics, %
reducing jitter \cite{RempeContactDynamics2020, Shimada2020PhysCapTOG, yuan2021simpoe,li2022dnd}, 
and 
(2) to discourage %
interpenetration
and encourage contact.
Since contact events are discontinuous, 
the \TODO{pipeline} is either not end-to-end trainable 
or %
trained with reinforcement learning \cite{peng2018deepmimic,yuan2021simpoe}.
\newtext{Xie \etal~\cite{Xie_2021_ICCV} propose differentiable physics-inspired objectives based on 
a soft contact penalty, while DiffPhy~\cite{gartner2022diffphy} uses a differentiable physics simulator~\cite{heiden2021neuralsim} during inference.}
\TODO{Both methods} %
apply the objectives in an \TODO{optimization scheme}, %
while \IPMAN is %
applied to
both optimization and regression. %
\optional{PhysCap~\cite{Shimada2020PhysCapTOG} 
considers a pose as \TODO{balanced}, when the \physCOM is projected within the \physBOS.}
Rempe \etal~\cite{RempeContactDynamics2020} impose PD control on 
the 
\newtext{pelvis}, 
which they treat as a \physCOM. %
Scott \etal~\cite{scott2020image} regress foot pressure from \twoD and \threeD 
joints %
for stability analysis but do not use it to improve 
\newtext{\HPS.}

All these methods 
use 
unrealistic %
bodies based on shape %
primitives.
Some require known body dimensions \cite{RempeContactDynamics2020,Shimada2020PhysCapTOG,yuan2021simpoe} %
while others estimate body scale %
\cite{Xie_2021_ICCV,li2022dnd}.
In contrast, \IPMAN computes \physCOM, \physCOP and \physBOS directly from the \smpl mesh.
Clever \etal~\cite{Clever2020bodiesRest} and Luo \etal~\cite{luo2021intelligent} estimate \threeD body pose but from pressure measurements, not from images. Their task is fundamentally different from ours.

%% file: sections_Paper/3_method.tex
\subsection{Preliminaries}
\label{subsec:preliminaries}

Given a color image, 
$\mathbf{I}$, we estimate the parameters of the camera and the \smpl body model \cite{SMPL:2015}.

\zheading{Body model}
\smpl maps pose, $\pose$, and shape, $\shape$, parameters to a %
\threeD mesh, $\boldsymbol{M}(\boldsymbol{\theta}, \boldsymbol{\beta})$. 
The pose parameters, $\pose \in \mathbb{R}^{24 \times 6}$, are %
rotations of 
\smpl's 24 joints 
\newtext{in a \sixD} representation \cite{zhou2019continuity}. 
The shape parameters, $\shape \in \mathbb{R}^{10}$, are %
the first 10 
PCA coefficients of \smpl's shape space. 
The generated mesh $\boldsymbol{M}(\boldsymbol{\theta}, \boldsymbol{\beta})$ consists of $N_V = 6890$ vertices, $\bs{V} \in \mathbb{R}^{N_V \times 3}$, 
and $N_F = 13776$ %
faces, $\bs{F} \in \mathbb{R}^{N_F \times 3 \times 3}$.

\newtext{Note that our 
regression method               (\nameMethodR, \cref{subsubsec:ipman-r})
uses \smpl, while our 
optimization method             (\nameMethodO, \cref{subsubsec:ipman-o}) uses
\smplx 
\cite{Pavlakos2019_smplifyx}, 
to match the models used by the baselines.
For simplicity of exposition, %
we refer to both models as \smpl when the distinction is not important.}

\zheading{Camera} 
For the regression-based \TODO{\nameMethodR}, we follow 
the 
standard convention 
\cite{Kanazawa2018_hmr, Kanazawa2019Learning3H, kolotouros2019spin} and use a weak perspective camera with a \twoD scale, $s$, translation, $\camtransl = (t_x^c, t_y^c)$, 
fixed camera rotation, $\camrot=\bs{I}_3$, and a fixed focal length $(f_x, f_y)$.
The root-relative body orientation $\bodyori$ is predicted by the neural network, but body translation stays fixed at $\bodytransl=\mathbf{0}$ as it is absorbed into the camera's translation. 

For the optimization-based \TODO{\nameMethodO}, we follow M\"uller \etal~\cite{Mueller:CVPR:2021} to use the full-perspective camera model and optimize the focal lengths $(f_x, f_y)$, camera rotation $\camrot$ and camera translation $\camtransl$. 
The principal point $(o_x, o_y)$ is the center of the input image. 
$\mathbf{K}$ is the intrinsic matrix storing focal lengths and the principal point.
We assume that the body rotation $\bodyori$ and translation $\bodytransl$ 
are absorbed into the camera parameters, 
thus, 
they %
stay fixed as $\bodyori=\bs{I}_3$ and $\bodytransl=\mathbf{0}$. 
Using the camera, we project a 3D point $\mathbf{X}\in \mathbb{R}^3$ to an image point $\mathbf{x}\in\mathbb{R}^2$ through $\mathbf{x} = \mathbf{K}(\mathbf{R}^c\mathbf{X}+\mathbf{t}^c)$.

\zheading{Ground plane and gravity-projection} 
We assume that the gravity direction is 
perpendicular to the ground plane 
in the world coordinate system. 
Thus, 
for any arbitrary point in \threeD space, $\bs{u} \in \mathbb{R}^3$, its \textit{gravity-projected} point, 
$\bs{u'} = g(\bs{u}) \in \mathbb{R}^3$, is the projection of $\bs{u}$ along the plane normal $\bs{n}$ onto the ground plane, and $g(.)$ is the projection operator. 
The function $h(\bs{u})$ returns the signed ``height'' of a %
point $\bs{u}$ 
with respect to 
the ground; \ie, the signed distance from $\bs{u}$ to the ground plane along the 
gravity direction, 
where 
$h(\bs{u}) < 0$ if $\bs{u}$ is below the ground and 
$h(\bs{u}) > 0$ if $\bs{u}$ is above it. %

\subsection{Stability Analysis}
\label{subsec:elements_of_stability_analysis}

We follow the biomechanics literature~\cite{hof2007equations, hof2008extrapolated, pai2003movement} and Scott \etal~\cite{scott2020image} to define three fundamental elements for stability analysis: 
We use %
the Newtonian definition for the ``\physExplainCOM'' (\physCOM); \ie, %
the mass-weighted average of 
particle positions. 
The ``\physExplainCOP'' (\physCOP) is 
the ground-reaction force's point of application. %
The ``\physExplainBOS'' (\physBOS) is %
the convex hull of all body-ground 
contacts. 
Below, 
we define intuitive-physics (\IP) %
terms using the inferred \physCOM and \physCOP. 
\physBOS is only used
for evaluation. 

\zheading{Body \physExplainCOM (\physCOM)} 
We introduce a novel \physCOM formulation that is fully differentiable and 
considers 
the per-part mass contributions, 
dubbed as \physCOMp; 
see \supmat for alternative \physCOM definitions.
To compute this, we first segment the template mesh %
into $N_P = 10$ parts $\meshParti \in \meshPartSET$; see~\reffig{figure:no_bos_loss}. 
We do this once offline, and keep the segmentation fixed during training and optimization. 
Assuming a shaped and posed \smpl body, the per-part volumes 
$\volumeParti$
are calculated by splitting the \smpl mesh into parts.

However, mesh splitting is a %
non-differentiable operation. 
Thus, it cannot be used for either training a regressor (\nameMethodR) or for optimization (\nameMethodO). 
Instead, we work 
with the full \smpl mesh 
and use differentiable \emph{``close-translate-fill''} operations for each body part on the fly. 
First, for each part $\meshPart$, we extract boundary vertices $ \mathcal{B}_{\meshPart}$ %
and add in the middle a 
\emph{virtual}
vertex $\bs{v}_g$, where 
$\bs{v}_g = \sum_{j\in\mathcal{B}_{\meshPart}}\bs{v}_j/|\mathcal{B}_{\meshPart}|$. 
Then, for the $\mathcal{B}_{\meshPart}$ and $\bs{v}_g$ vertices, we add virtual faces to \emph{``close''} $\meshPart$ and make it \emph{watertight}. 
Next, we \textit{``translate"} $\meshPart$
such that the part centroid $\mathbf{c}_{\meshPart} = \sum_{j \in \meshPart} \bs{v}_j / | P |$ 
is at the origin. 
Finally, we \textit{``fill''} the centered $\meshPart$ with tetrahedrons by connecting the origin with each face 
vertex. 
Then, the part volume, 
$\volumePart$, 
is the sum of all 
tetrahedron volumes  
\cite{zhang2001efficient}. 

To create a uniform distribution of surface vertices,
we  uniformly sample $N_U = 20000$ %
surface points $\bs{V}_U \in \mathbb{R}^{N_U \times 3}$ on the template 
\smpl mesh using the Triangle Point Picking method~\cite{weiss_tpp}. Given $\bs{V}_U$ and the template \smpl mesh vertices $\bs{V}_{T}$, we \newcamready{follow \cite{Mueller:CVPR:2021}, and} analytically compute a sparse linear regressor $\mathbf{W} \in \mathbb{R}^{N_U \times N_V}$ such that $\bs{V}_U = \mathbf{W}\bs{V}_{T}$. During training and optimization, given an arbitrary shaped and posed mesh with vertices $\bs{V}$, we obtain uniformly-sampled mesh surface points as $\bs{V}_U=\mathbf{W}\bs{V}$. Each surface point, $v_i$, is assigned to the body part, $\meshPartvi$, corresponding to the face, $\bs{F}_{v_i}$, it was sampled from. 

Finally, the part-weighted \physCOMp is computed as a volume-weighted mean of the mesh surface points:
\begin{equation}
    \bar{\mathbf{m}} = \frac{\sum_{i=1}^{N_U} \volumePartvi v_i}{\sum_{i=1}^{N_U} \volumePartvi}    \text{,}
\end{equation}%
where 
$\volumePartvi$ is the volume of the part $\meshPartvi \in \meshPartSET$ to which $v_i$ is assigned. 
This formulation is fully differentiable and can be employed with any existing \threeD \hps estimation method.

\newcamready{Note that computing \physCOM (or volume) from uniformly sampled surface points does not work (see \supmat) because it assumes that %
mass, $M$, is proportional to surface area, $S$. 
Instead, our \physCOMp computes mass from volume, $\mathcal{V}$, via the standard density 
equation, 
$M=\rho\mathcal{V}$, %
while our \emph{close-translate-fill} operation computes the volume of deformable bodies 
in an efficient and differentiable manner.}

\zheading{Center of Pressure (\physCOP)} 
Recovering a pressure heatmap from an image without 
using hardware, such as pressure sensors, 
is a highly ill-posed problem.
However, %
stability analysis requires knowledge of the pressure exerted on the human body by the supporting surfaces, like the ground. 
Going beyond binary contact, Rogez \etal~\cite{Rogez2015everyday} estimate \threeD forces by detecting intersecting vertices between %
hand and object meshes. 
Clever \etal~\cite{Clever2020bodiesRest} recover %
pressure maps by allowing articulated
body models to deform a soft pressure-sensing virtual mattress in a physics simulation. 

In contrast,
we observe that, while real bodies 
interacting with rigid objects (\eg, the floor) deform under contact, 
\smpl does not model 
such 
\newtext{soft-tissue} 
deformations. 
\newtext{Thus}, 
the body mesh penetrates the contacting object surface and the amount of penetration can be %
a proxy for pressure;
a deeper penetration implies higher pressure. 
With the height $h(v_i)$ (see \cref{subsec:preliminaries}) of a 
mesh surface point $v_i$ 
with respect to the ground plane $\Pi$, \newtext{we define a \emph{pressure field}} to compute the  
per-point 
pressure $\rho_i$ as: 
\begin{eqnarray}
    \rho_i &=& 
    \begin{cases}
        1 - \alpha h(v_i) & \text{if~} h(v_i) <    0  \text{,} \\
        e^{-\gamma  h(v_i)} & \text{if~}  h(v_i) \geq 0  \text{,}
    \end{cases} 
\end{eqnarray}
where $\alpha$ and $\gamma$ are 
scalar hyperparameters set empirically. \newcamready{We approximate soft tissue via a ``spring'' model and ``penetrating'' pressure field using Hooke's Law.} 
Some pressure is also assigned to points above the ground to allow tolerance for footwear, but this decays quickly. 
Finally, we compute the \physCOP, $\mathbf{\bar{s}}$, as
\begin{eqnarray}
\mathbf{\bar{s}} &=& \frac{\sum_{i=1}^{N_U}\rho_i v_i}{\sum_{i=1}^{N_U}\rho_i} \text{.}
\end{eqnarray}
Again, note that this term is fully differentiable. %

\zheading{Base of Support (\physBOS)} 
In biomechanics~\cite{HOFdynamic20051, winter1995abc}, \physBOS is defined as the ``supporting area'' or the possible range of the \physCOP on the supporting surface. 
Here, 
we define \physBOS as the convex hull~\cite{rockafellar2015convex} of all gravity-projected body-ground contact points. 
In detail, we first determine all 
such contacts 
by selecting the set of mesh surface points $v_i$ close to the ground, %
and then gravity-project them onto the ground %
to obtain $C = \{g(v_i) \; \bigr |\; |h(v_i)| < \tau \}$. The \physBOS is then defined as the convex hull $\mathcal{C}$ of $C$.

\subsection{Intuitive-Physics Losses}
\label{subsec:ip_losses}

\zheading{Stability loss} 
The \emph{``inverted pendulum''} model of human balance~\cite{winter1995abc, winter1995human} considers the relationship between the \physCOM and \physBOS to determine stability. Simply put, for a given shape and pose, if the body \physCOM, projected on the gravity-aligned ground plane, lies within the \physBOS, the pose is considered \emph{stable}. While this definition of stability is useful 
for evaluation, using it 
in a loss or energy function for \threeD \hps estimation results in sparse gradients (see \supmat). Instead, we define the stability criterion as: 
\begin{eqnarray}
    \mathcal{L}_{\text{stability}} &=& \| g(\bar{\mathbf{m}}) \; - \; g(\bar{\mathbf{s}}) \|_2    \text{,}
    \label{eq:loss_stability}
\end{eqnarray}
where $g(\bar{\mathbf{m}})$ and $g(\bar{\mathbf{s}})$ are the gravity-projected \physCOM and \physCOP, respectively. 

\zheading{Ground contact loss}
As shown in \reffig{figure:teaser_new}, \threeD \hps methods 
minimize 
the \twoD joint reprojection error and do not consider the 
plausibility of 
body-ground contact. 
Ignoring this %
can result in interpenetrating %
or hovering meshes. 
Inspired by 
self-contact losses \cite{fieraru2021remips,Mueller:CVPR:2021} and 
hand-object contact losses \cite{hampali2020honnotate,hasson2019obman}, 
we define two ground losses, namely  pushing, $\mathcal{L}_{\text{push}}$, and pulling, $\mathcal{L}_{\text{pull}}$, that take into account the height, $h(v_i)$, of a vertex, $v_i$, 
with respect to the ground plane. 
For $h(v_i)<0$, \ie, for vertices under the ground plane, $\mathcal{L}_{\text{push}}$ discourages body-ground penetrations. 
For $h(v_i)\geq0$, \ie, for hovering meshes, $\mathcal{L}_{\text{pull}}$ encourages the vertices that lie close to the ground to ``snap'' into contact with it. 
\newtext{Note that the losses are non-conflicting as they act on disjoint sets of vertices.} 
Then, the ground contact loss is:
\begin{eqnarray}
    \mathcal{L}_{\text{ground}} &=& \mathcal{L}_{\text{pull}} + \mathcal{L}_{\text{push}} \text{, with } \\
    \mathcal{L}_{\text{pull}}   &=& \alpha_1\tanh{(\frac{h(v_i)}{\alpha_2})^2} \quad \text{if~}h(v_i)\geq0 \text{, and } \\
    \mathcal{L}_{\text{push}}   &=& \beta_1\tanh{(\frac{h(v_i)}{\beta_2})^2} \quad \text{if~}h(v_i)<0 \text{.}
\end{eqnarray}

\subsection{\nameMethod}

We use our new \IP losses for two tasks:
(1)
We extend \hmr~\cite{Kanazawa2018_hmr} to develop \nameMethodR, a regression-based \hps method. 
(2)
We extend \smplifyXMC~\cite{Mueller:CVPR:2021} to develop \nameMethodO, an optimization-based method. 
Note that 
\nameMethodO uses a reference ground plane, while 
\nameMethodR uses the ground plane only for training but not at test time. \newcamready{It leverages the \emph{known} ground in \threeD datasets, and thus, does not require additional data beyond past \hps methods.}

\subsubsection{\nameMethodR} 
\label{subsubsec:ipman-r}

\begin{figure}[t]
\vspace{-0.6 em}
\centerline{
\includegraphics[width=\linewidth]{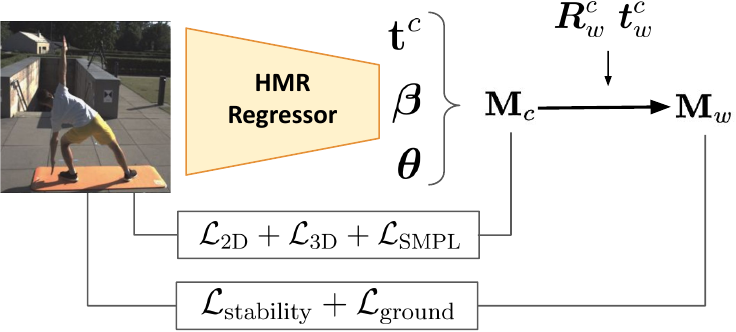}
}
\vspace{-0.5 em}
\caption{
    \nameMethodR architecture. 
    First, the \hmr regressor estimates camera translation and \smpl parameters for an input image. 
    These parameters are used to generate the \smpl mesh in the camera frame, $\bs{M}_c$. 
    To transform the mesh from camera into world coordinates ($\bs{M}_c \rightarrow \bs{M}_w$), \nameMethodR uses the ground-truth 
    camera rotation, $\bs{R}^c_w$, and 
    translation, $\bs{t}^c_w$. 
    The \IP losses, $\mathcal{L}_{\text{ground}}$ and $\mathcal{L}_{\text{stability}}$, are applied on the mesh in the world coordinate system. 
}
\label{figure:ipmanr_architecture}
\end{figure}

Most \HPS methods are trained with
a mix of direct supervision using \threeD datasets~\cite{ionescupapavaetal2014, Mehta2017Monocular3H, vonmarcard_eccv_2018_3dpw} and \twoD reprojection losses using image datasets~\cite{Lin2014MicrosoftCC, Johnson10, andriluka14cvpr}.
The \threeD losses, however, are calculated in the camera frame, %
ignoring scene information and physics. 
\nameMethodR extends \hmr~\cite{Kanazawa2018_hmr}
with our intuitive-physics terms; %
\newtext{see \cref{figure:ipmanr_architecture} for the architecture.} %
For training, we use the \textit{known} camera coordinates and the world ground plane in \threeD datasets.

As described in \refsec{subsec:preliminaries} (paragraph ``Camera''), \hmr infers the camera translation, $\mathbf{t}^c$, and \smpl parameters, $\pose$ and $\shape$, in the camera coordinates assuming 
$\camrot=\bs{I}_3$ and 
$\bodytransl=\mathbf{0}$. 
\newtext{Ground truth \threeD joints and \smpl parameters are used to supervise the inferred mesh $\bs{M}_c$ in the \emph{camera frame}. However, \threeD datasets also provide the ground, albeit in the world frame. To leverage the known ground, we}
transform the predicted body orientation, $\mathbf{R}^b$, to world coordinates using the ground-truth camera rotation, $\mathbf{R}^c_w$, as $\mathbf{R}^b_w =\mathbf{R}^{c\top}_w\mathbf{R}^b $. 
Then, we compute the body translation in world coordinates as $\mathbf{t}^b_w = - \mathbf{t}^c + \mathbf{t}^c_w$. With the predicted mesh and ground plane in world coordinates, we add the \IP terms, $\mathcal{L}_\text{stability}$ and $\mathcal{L}_\text{ground}$, for \HPS training as follows: 
\begin{align}
    \mathcal{L}_\text{\nameMethodR}(\bs{\theta}, \bs{\beta}, \camtransl) = %
    & \lambda_{\twoD}\mathcal{L}_{\twoD} + \lambda_{\threeD}\mathcal{L}_{\threeD} + \lambda_{\text{SMPL}}\mathcal{L_{\text{SMPL}}} + \nonumber \\
    & \lambda_{\text{s}}\mathcal{L}_\text{stability} + \lambda_{\text{g}}\mathcal{L}_\text{ground} \text{,}
\end{align}%
where $\lambda_{\text{s}}$ and $\lambda_{\text{g}}$ are the weights for the respective \IPterms.
For training (data augmentation, hyperparameters, 
\etc), we follow Kolotouros \etal~\cite{kolotouros2019spin}; 
for more details see \supmat

\subsubsection{\nameMethodO} 
\label{subsubsec:ipman-o}

To fit \smplx  
to \twoD image keypoints, %
\TODO{\smplifyXMC \cite{Mueller:CVPR:2021} initializes the fitting process by exploiting the self-contact and global-orientation of a known/presented \threeD mesh.} 
We posit that the presented pose contains further information, such as stability, pressure and contact with the ground-plane. \nameMethodO uses this insight 
to apply stability and ground contact losses. 
The \nameMethodO objective is: %
\begin{align}
    E_\text{\nameMethodO}(\bs{\beta}, \bs{\theta}, \bs{\Phi}) = 
    &
    E_{J2D} + 
    \lambda_{\beta} E_{\beta} + \lambda_{\theta_h} E_{\theta_h} + \nonumber \\ &\lambda_{\tilde{\theta_b}} E_{\tilde{\theta_b}} +  \lambda_{\tilde{C}} E_{\tilde{C}} + \nonumber \\
    &\lambda_{s} E_{\text{stability}} + \lambda_{g} E_{\text{ground}} \text{.}
\end{align}
$\bs{\Phi}$ denotes the camera parameters: rotation $\camrot$, translation $\camtransl$, and focal length, $(f_x, f_y)$. 
$E_{J2D}$ is a \twoD joint loss, $E_{\beta}$ and $E_{\theta_h}$ are $L_2$ body shape and hand pose priors. $E_{\tilde{\theta_b}}$ and $E_{\tilde{C}}$ are pose and contact terms 
\wrt 
the presented \threeD pose and contact (see \cite{Mueller:CVPR:2021} for details). $E_{S}$ and $E_G$ are the stability and ground contact losses 
from 
\cref{subsec:ip_losses}. 
Since the estimated mesh is in the same coordinate system as the presented mesh and the ground-plane, we directly apply \IP losses without any transformations. 
For details see \supmat

%% file: sections_Paper/4_experiments.tex
\subsection{Training and Evaluation Datasets}

\qheading{Human3.6M \cite{ionescupapavaetal2014}}
A dataset of \threeD human keypoints and \rgb images. 
The poses are limited in terms of challenging physics, focusing on common activities like walking, discussing, smoking, or taking photos. 

\qheading{\rich~\cite{RICH}}
A dataset of videos with accurate marker-less motion-captured \threeD bodies and \threeD scans of scenes. 
The images are more natural than %
Human3.6M and Fit3D \cite{Fieraru_2021_CVPR}.
We consider sequences %
with 
meaningful body-ground interaction.
For the list of sequences, see \supmat

\qheading{Other datasets}
Similar to \cite{kolotouros2019spin}, for training we use 3D keypoints  from MPI-INF-3DHP~\cite{Mehta2017Monocular3H} and 2D keypoints  from image datasets such as COCO~\cite{Lin2014MicrosoftCC}, MPII~\cite{andriluka14cvpr} and LSP~\cite{Johnson10}.  

\input{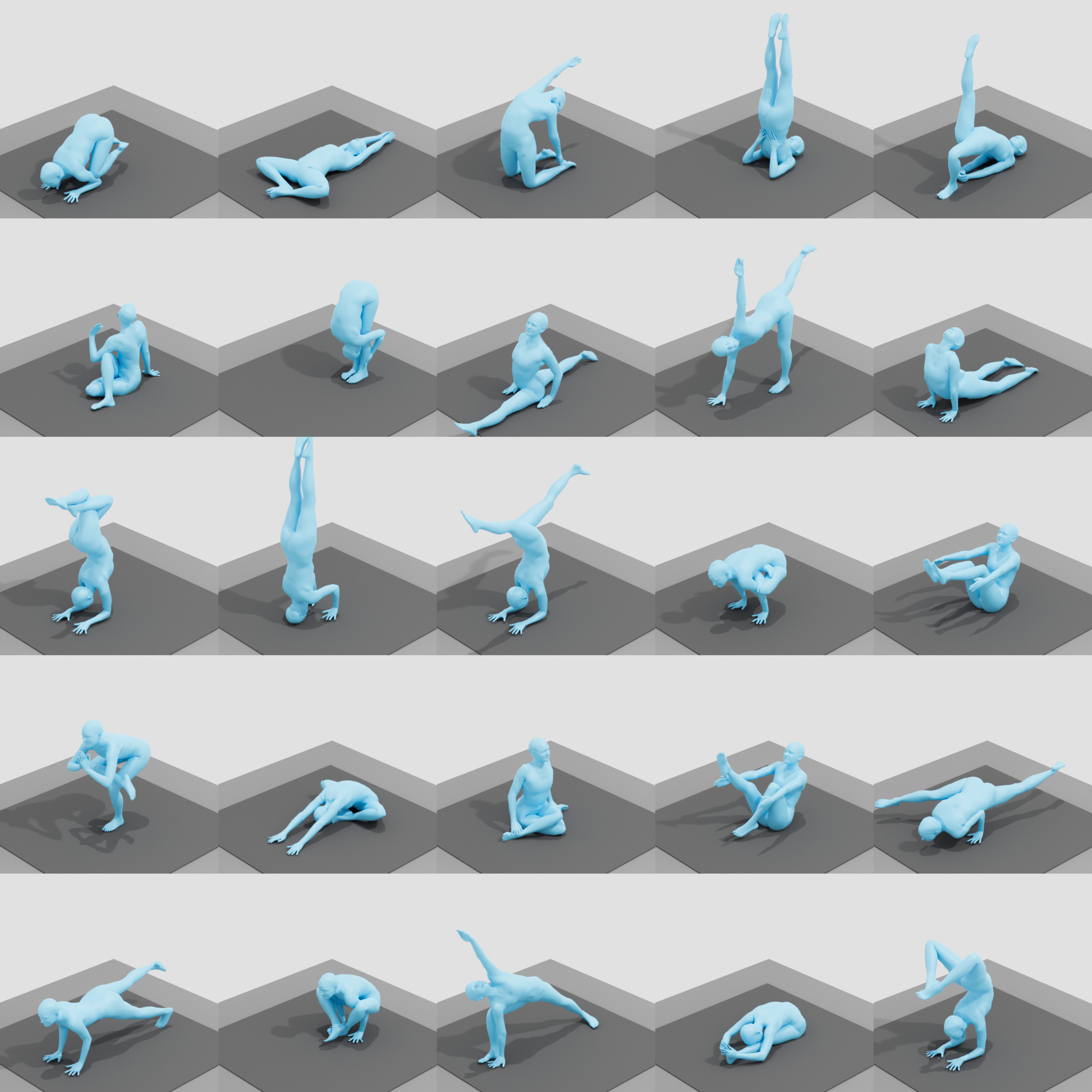}

\subsubsection{\nameYOGIDataBig (\nameYOGIData) \newcamready{Dataset}}

We capture a trained Yoga professional in 200 highly complex poses (see \cref{figure:ipman:moyo_montage}) using a synchronized \mocap system, pressure mat, and a multi-view RGB video system with 8 static, calibrated cameras; 
for details see \supmat 
\newtext{The dataset  contains $\sim1.75$M RGB frames in 4K resolution with ground-truth SMPL-X~\cite{Pavlakos2019_smplifyx}, pressure and \physCOM. Compared to the Fit3D \cite{Fieraru_2021_CVPR} and PosePrior \cite{Akhter:CVPR:2015} datasets, \nameYOGIData is more challenging; it has extreme poses, strong self-occlusion, and significant body-ground and self-contact.}

\subsection{Evaluation Metrics}

We use standard \threeD \HPS metrics: 
The 
Mean Per-Joint Position Error (\mpjpe), its Procrustes Aligned version (\pampjpe), and  
the 
Per-Vertex Error (\pve)~\cite{patel2021agora}. 

\qheading{\physBOS Error (\physBOSE)}
To evaluate stability, we propose a new metric called \physBOS Error (\physBOSE). Following the definition of stability 
(\cref{eq:loss_stability}) 
we define:
\begin{eqnarray}
    \physBOSE &=& 
    \begin{cases}
        1 & g(\bar{\mathbf{m}}) \in \mathcal{C}(C) \\
        0 & g(\bar{\mathbf{m}}) \notin \mathcal{C}(C) \\
    \end{cases} 
\end{eqnarray}
where $\mathcal{C}(C)$ is the convex hull of the gravity-projected contact vertices for $\tau = 10$ cm. For efficiency reasons, we formulate this computation as the solution of a convex system via interior point linear programming~\cite{Andersen2000TheMI}; see \supmat

\subsection{\nameMethod Evaluation}

\zheading{\nameMethodR}
We evaluate our regressor, \nameMethodR, on \rich and \humanthreesix and summarize our results in \reftab{table:sota_regression_hmr}. We refer to our regression baseline as $\text{\hmr}^*$ which is \hmr trained on the same datasets as \nameMethodR. Since we train with paired 3D datasets, we do not use \hmr's discriminator during training. Both \IPterms individually improve upon the baseline method. Their joint use, however, shows the largest improvement. For example, on \rich the \mpjpe improves by 3.5mm and the \pve by 2.5mm. It is particularly interesting that \nameMethodR  improves upon the baseline on \humanthreesix, a dataset with largely dynamic poses and little body-ground contact. We also significantly outperform ($\sim12\%$) the MPJPE of optimization approaches that use the ground plane,  
Zou et al.~\cite{zou2020reducing} (69.9 mm) and 
Zanfir et al.~\cite{zanfir2018monocular} (69.0 mm), on \humanthreesix.
\newtext{Some video-based methods ~\cite{yuan2021simpoe, li2022dnd} achieve better \mpjpe ($56.7$ and $52.5$ resp.) on \humanthreesix. However, they initialize with a stronger kinematic predictor~\cite{VIBE:CVPR:2020, li2021hybrik} and require video frames as input. Further, they use heuristics to estimate body weight and non-physical residual forces to \emph{correct} for contact estimation errors. In contrast, \nameMethod is a  single-frame method, models complex full-body pressure and does not rely on approximate body weight to compute \physCOM.}
Qualitatively, \cref{figure:ipman_qual_result_both}~(top) shows that \nameMethodR's reconstructions are more stable and contain physically-plausible body-ground contact.  
\newtext{While \hmr is not SOTA, it is simple, isolating the benefits of our new IP formulation.
These terms can also be added to methods with more modern backbones and architectures.
}

\input{TEX_tables/sota_regr_hmr}

\input{TEX_figures/ipman_qual_results_combined}

\zheading{\nameMethodO}
Our optimization method, \nameMethodO, also improves upon the baseline optimization method, \smplifyXMC, on all evaluation metrics (see \cref{table:sota_optimization}). We note that adding $L_\text{stability}$ independently improves the \pve, but not joint metrics (\pampjpe, \mpjpe) and \physBOSE.
This can be explained by the dependence of our \IPterms on the relative position of the mesh surface to the ground-plane.
Since joint metrics do not 
\SUBSPRINT{capture surfaces}, 
they may get worse.
Similar trends on joint metrics have been reported in the context of hand-object contact~\cite{hasson2019obman, Tzionas2016CapturingHI} and body-scene contact~\cite{Hassan2019prox}. We show qualitative results in \cref{figure:ipman_qual_result_both} (bottom). While both \smplifyXMC~\cite{Mueller:CVPR:2021} and \nameMethodO achieve similar image projections, another view reveals that our results are more stable and physically plausible 
\SUBSPRINT{\wrt the ground.} 

\input{images/method/gt_vs_pred_pressuremaps}

\subsection{Pressure, \physCOP and \physCOM Evaluation}

\newtext{We evaluate our estimated pressure, \physCOP and \physCOM against the \nameYOGIData ground truth. %
For pressure evaluation, we measure Intersection-over-Union (IoU) between our estimated  and ground-truth pressure heatmaps. 
We also compute the \physCOP error as the Euclidean distance between estimated and ground-truth \physCOP. 
We obtain an IoU of $0.32$ and a \physCOP error of $57.3$ mm.
\Cref{figure:ipman:gt_vs_pred_pressuremaps} 
shows a qualitative visualization of the estimated pressure compared to the ground truth. 
For \physCOM evaluation, 
we find a $53.3$ mm difference between our \physCOMp and 
the \physCOM computed by the commercial software, Vicon Plug-in Gait. 
Unlike Vicon's estimate,
our \physCOMp does not require anthropometric measurements and takes into account the full \threeD body shape. 
For details about the evaluation protocol \newcamready{and comparisons with alternative \physCOM formulations}, see \supmat
}

\noindent \newcamready{\textbf{Physics Simulation.} To evaluate stability, we run a post-hoc physics simulation in ``Bullet"~\cite{pyBullet} and measure the displacement of the estimated meshes; a small displacement denotes a stable pose. \nameMethodO produces $14.8\%$ more stable bodies than the baseline~\cite{Mueller:CVPR:2021}; for details see \supmat}  

\input{TEX_tables/sota_opti}

%% file: images/experiments/moyo_montage.tex
\begin{figure*}[ht]
\centerline{
\includegraphics[width=\linewidth]{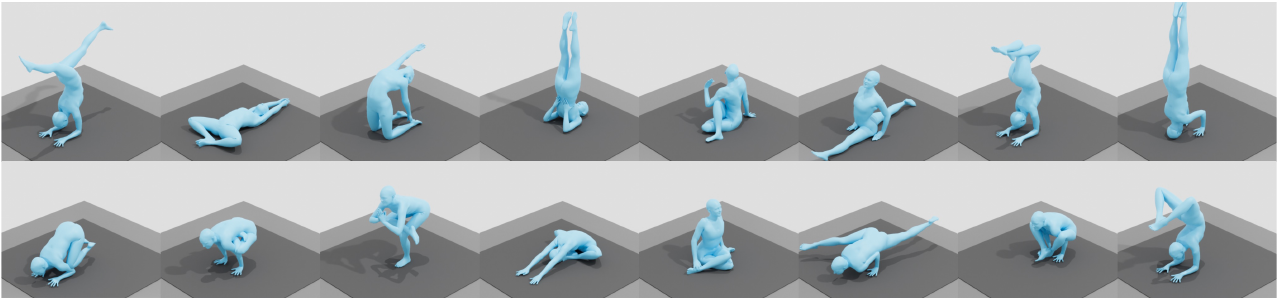}
}
\vspace{-0.5 em}
\caption{
        \SUBSPRINT{Representative} 
        examples illustrating the variation and complexity of 3D pose and body-ground contact in our \SUBSPRINT{new} \nameYOGIData dataset.}
\label{figure:ipman:moyo_montage}
\vspace{+0.5 em}
\end{figure*}

%% file: TEX_tables/sota_regr_hmr.tex
\begin{table}[t]
    \centering
    \small
    \rowcolors{3}{}{lightgray}
    \renewcommand{\arraystretch}{1.2} %
    \setlength{\tabcolsep}{4pt}
    \begin{adjustbox}{width=1\linewidth}
    \begin{tabular}{@{}l|l c c c | l c @{}}
    \Xhline{3\arrayrulewidth}
   
    \multirow{2}{*}{\textbf{Method}} & \multicolumn{4}{c|}{\textbf{\rich}} & \multicolumn{2}{c}{\textbf{Human3.6M}}  \\
    
     & \textbf{MPJPE $\downarrow$} & \textbf{PAMPJPE $\downarrow$} & \textbf{PVE $\downarrow$} & \textbf{\physBOSE (\%) $\uparrow$} & \textbf{MPJPE $\downarrow$} & \textbf{PAMPJPE $\downarrow$}\\
    \hline
    PhysCap~\cite{Shimada2020PhysCapTOG} & - & - & - & - & 113.0 & 68.9 \\
    DiffPhy~\cite{gartner2022diffphy} & - & - & - & - & 81.7 & 55.6 \\
    Zou et al.~\cite{zou2020reducing} & - & - & - & - & 69.9 & - \\
    Xie \etal~\cite{Xie_2021_ICCV} & - & - & - & - & 68.1 & - \\ 
    VIBE~\cite{VIBE:CVPR:2020} & - & - & - & - & 61.3 & 43.1 \\
    Simpoe~\cite{yuan2021simpoe} & - & - & - & - & 56.7 & 41.6 \\
    D\&D~\cite{li2022dnd} & - & - & - & - & \textbf{52.5} & \textbf{35.5} \\
    \hline
    \hmr~\cite{Kanazawa2018_hmr} & - & - & - & - & 88.0 & 56.8 \\
    Zanfir~\etal~\cite{zanfir2018monocular} & - & - & - & - & 69.0 & - \\
    SPIN~\cite{kolotouros2019spin} & 112.2 & 71.5 & 129.5 & 54.7 & 62.3 & 41.9  \\
    PARE~\cite{Kocabas2021pare} & 107.0 & 73.1 & 125.0 & 74.4 & - & -  \\
    CLIFF~\cite{li2022cliff} & 107.0 & 67.2 & 122.3 & 67.6 & 81.4 & 52.1\\
    \hline
    \multicolumn{7}{c}{Finetuning on Human3.6M} \\
    \hline
    $\text{\hmr}^*$~\cite{Kanazawa2018_hmr} & - & - & - & - & 62.1 & 41.6 \\ %
    \nameMethodR (Ours) & - & - & - & - & \textbf{60.7 (-1.4)} & \textbf{41.1 (-0.5)} \\ %
    \hline
    \multicolumn{7}{c}{Finetuning on all datasets} \\
    \hline
    $\text{\hmr}^*$~\cite{Kanazawa2018_hmr} & 82.5 & 48.3 & 92.4 & 62.0 & 61.6 & 41.9   \\ %
    $\text{\hmr}^*$~\cite{Kanazawa2018_hmr}$+\mathcal{L}\text{ground}$ & 80.9 & 47.8 & 89.9 & 66.5 & 61.9 & 41.8   \\ %
    $\text{\hmr}^*$~\cite{Kanazawa2018_hmr}$+\mathcal{L}\text{stability}$ & 81.0 & \textbf{47.5 (-0.8)} & 90.8 & 69.6 & 61.2  & 41.9  \\ %
    \nameMethodR (Ours) & \textbf{79.0 (-3.5)} & 47.6  & \textbf{89.9 (-2.5)} & \textbf{71.2 (+9.2)} & \textbf{60.6 (-1.0)} & \textbf{41.8 (-0.1)}  \\ %
    \hline
    \Xhline{3\arrayrulewidth}
    \end{tabular}
    \end{adjustbox}
\vspace*{-0.5 em}
    \caption{Top to Bottom: Comparisons with \newtext{video-based and single-frame regression methods}. \nameMethodR outperforms the single-frame baselines across all benchmarks. * indicates training hyperparameters and datasets are identical to \nameMethodR. All units are in mm except \physBOSE. 
    \SUBSPRINT{Bold denotes best results (per category), and parentheses show improvement over the baseline. \faSearch~\textbf{Zoom in}} 
    }
    \label{table:sota_regression_hmr}
\vspace*{-0.5 em}
\end{table}

%% file: TEX_figures/ipman_qual_results_combined.tex
\begin{figure*}[ht]
\vspace{-0.5 em}
\centerline{
\includegraphics[width=\linewidth]{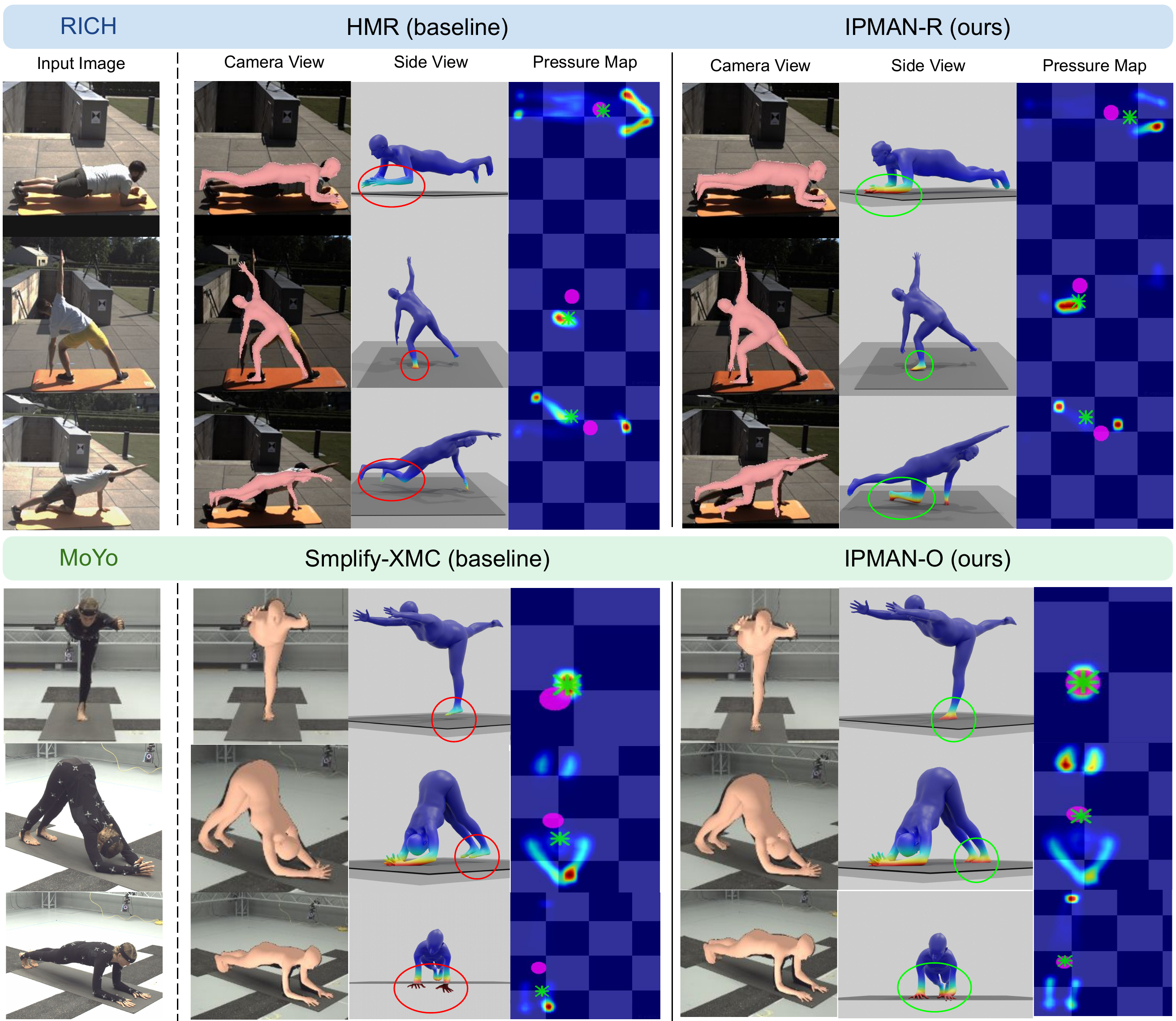}
}
\vspace{-0.5 em}
\caption{Qualitative evaluation of \nameMethodR and \nameMethodO on the \rich and \nameYOGIData datasets. 
The first column shows the input images of a subject doing various sports poses. 
The second and third block of columns show 
\SUBSPRINT{the baseline's and our} 
results, respectively. 
In each block, the first image shows the estimated mesh overlayed on the image \SUBSPRINT{(camera view)}, 
\newtext{the second image shows the estimated mesh in the world frame \SUBSPRINT{(side view)}, and the last image shows the \newtext{estimated} pressure map with the \physCOM (in \textcolor{magenta}{pink}) and the \physCOP (in \textcolor{green}{green}).}
}
\label{figure:ipman_qual_result_both}
\end{figure*}

%% file: images/method/gt_vs_pred_pressuremaps.tex
\begin{figure}[t]
\centerline{
\includegraphics[width=\linewidth]{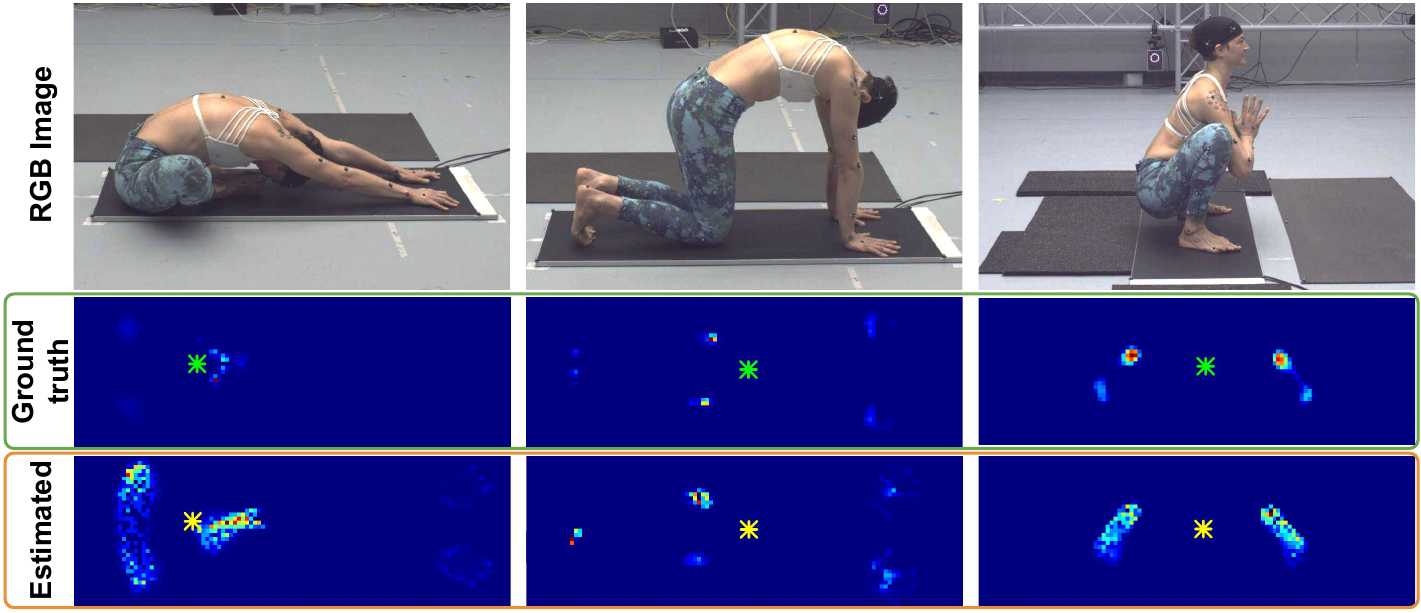}}
\vspace{-0.1in}
\caption{
            \newtext{
                Qualitative comparison of estimated vs the ground-truth pressure. 
                The ground-truth \physCOP is shown in \textcolor{teal}{green} and the estimated \physCOP is shown in \textcolor{orange}{yellow}. 
                Pressure heatmap colors as per \cref{figure:no_bos_loss}.
            }
}
\label{figure:ipman:gt_vs_pred_pressuremaps}
\end{figure}

%% file: TEX_tables/sota_opti.tex
\begin{table}[t]
    \centering
    \small
    \rowcolors{3}{}{lightgray}
    \renewcommand{\arraystretch}{1.2} %
    \setlength{\tabcolsep}{4pt}
    \begin{adjustbox}{width=1\linewidth}
    \begin{tabular}{@{}l|l c c c @{}}
    \Xhline{3\arrayrulewidth}
   
    \multirow{2}{*}{\textbf{Method}}  & \multicolumn{4}{c}{\textbf{\nameYOGIData}}  \\
    
     & \textbf{MPJPE $\downarrow$} & \textbf{PAMPJPE $\downarrow$} & \textbf{PVE $\downarrow$} & \textbf{\physBOSE (\%) $\uparrow$} \\
    \hline
    \smplifyXMC~\cite{Mueller:CVPR:2021}  & 75.3 & 36.5 & 16.8 & 98.0   \\
    \smplifyXMC~\cite{Mueller:CVPR:2021}$+\mathcal{L}\text{ground}$ & 73.3 & 36.2 & 14.5 & 98.2 \\
    \smplifyXMC~\cite{Mueller:CVPR:2021}$+\mathcal{L}\text{stability}$ & 88.5 & 38.6 & 15.3 & 97.8 \\
    \nameMethodO (Ours) & \textbf{71.9 (-3.4)} & \textbf{34.3 (-2.2)} & \textbf{11.4 (-5.4)} & \textbf{98.6 (+0.5)} \\  %
    \hline
    \Xhline{3\arrayrulewidth}
    \end{tabular}
    \end{adjustbox}
\vspace*{-0.5 em}
    \caption{
            Evaluation of \nameMethodO and 
            \smplifyXMC \cite{Mueller:CVPR:2021}
            (optimization-based) 
            on \nameYOGIData.
            \SUBSPRINT{Bold shows the best performance, and parentheses show the improvement over \smplifyXMC.} 
    }
    \label{table:sota_optimization}
\vspace*{-0.5 em}
\end{table}

%% file: sections_Paper/5_conclusion.tex
Existing \threeD \hps estimation methods recover SMPL meshes that align well with the input image, but are often physically implausible. 
To address this, we propose \nameMethod, which incorporates \emph{intuitive-physics} in \threeD \hps estimation. Our \IPterms encourage  stable poses, promote realistic floor support, and reduce body-floor penetration. The \IPterms exploit the interaction between the body \physCOM, \physCOP, and \physBOS~-- key elements used in stability analysis. To calculate the \physCOM of SMPL meshes,  \nameMethod uses on a novel formulation that takes part-specific mass contributions into account. Additionally, \nameMethod  estimates proxy \emph{pressure} maps directly from images, which is useful in computing \physCOP. \nameMethod is simple, differentiable, and compatible with both regression and optimization methods. 
\nameMethod goes beyond previous physics-based methods to reason about arbitrary full-body contact with the ground.
We show that \nameMethod improves both regression and optimization baselines across all metrics on 
\newtext{existing datasets and MoYo.}
\SUBSPRINT{%
\nameYOGIData uniquely comprises synchronized multi-view video, \smplX bodies in complex poses, 
and measurements for pressure maps and body \physCOM.}
Qualitative results show the effectiveness of \nameMethod in recovering physically plausible meshes. 
 
While \nameMethod addresses body-floor contact, future work should incorporate general body-scene contact and diverse supporting surfaces \newtext{by integrating 3D scene reconstruction}. In this work, the proposed \IPterms are designed to help static poses and we show that they do not hurt dynamic poses. However, the large body of biomechanical literature analyzing dynamic poses could be leveraged for activities like walking, jogging, running, \etc. \newtext{It would be interesting to extend \nameMethod beyond single-person scenarios by exploiting the various physical constraints offered by multiple subjects. }

%% file: sections_Paper/6_acknowledgements.tex
{\small
	\qheading{Acknowledgements} 
	We thank T.~Alexiadis, G.~Becherini, T.~McConnell, C.~Gallatz, M.~H\"{o}schle, S.~Polikovsky, C.~Mendoza, Y.~Fincan, L.~Sanchez and M.~Safroshkin for the \nameYOGIData data,  
	J.~Tesch, N.~Athanasiou, Z.~Fang, V.~Choutas and all of Perceiving Systems for fruitful discussions. 
	This work is funded by the International Max Planck Research School for Intelligent Systems (IMPRS-IS) and in part by
	the German Federal Ministry of Education and Research (BMBF), T{\"u}bingen AI Center, FKZ: 01IS18039B.}

\noindent \small \textbf{Disclosure.} \href{https://files.is.tue.mpg.de/black/CoI_CVPR_2023.txt}{https://files.is.tue.mpg.de/black/CoI\_CVPR\_2023.txt} 

%% file: supplementary/sections_Supmat/moyo_details.tex
\section{\nameYOGIDataBig Dataset (\nameYOGIData)}
\label{supmat:moyo_details}

\begin{figure}[ht]
\centerline{
\includegraphics[width=\linewidth]{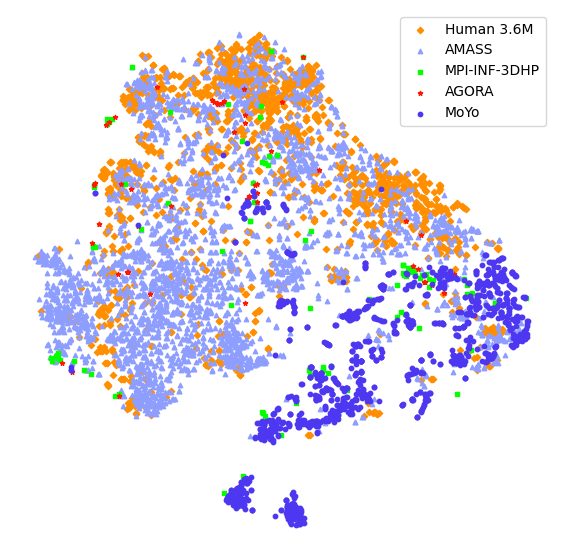}
}
\caption{The distribution of poses in \nameYOGIData
and existing \mocap datasets are visualized after T-SNE dimension reduction.}
\label{figure:ipman:moyo_tsne}
\end{figure}

\newtext{We capture a trained yoga professional in a \mocap studio with 54 Vicon Vantage V16 infrared cameras capable of tracking body markers as small as 3mm in diameter. The Vicon system was synchronized with 8 RGB cameras recording at 4112x3008 resolution and a Zebris FDM pressure measurement mat. The pressure mat offers a sensor resolution of 1.4sensors/cm$^2$ and can capture pressure in 10-1200 kPa range. Ground-truth \smplx~\cite{Pavlakos2019_smplifyx} parameters are recovered from the \mocap data using MoSh++~\cite{AMASS_2019}. A total of 200 yoga sequences were recorded at 30fps. The yoga poses we selected include all poses in the Yoga-82 dataset~\cite{verma2020yoga} as well as their variations. The T-SNE~\cite{JMLR:v9:vandermaaten08a} plot in \cref{figure:ipman:moyo_tsne} shows that the poses contained in \nameYOGIData are highly diverse and cover areas in the space of  human poses not well represented in existing datasets~\cite{ionescupapavaetal2014, AMASS_2019, Mehta2017Monocular3H, patel2021agora}}.

\newtext{To compute a reference  \physCOM, we use the commercially available tool, \emph{Plug-in Gait (PiG)} from Vicon. PiG requires
a-priori 
known 
anthropometric measurements (\eg height, weight, shoulder offset, knee width, \etc) and computes: 
(1) bone joints from a known marker topology, %
(2) per-bone mass as a proportion of %
body mass, 
(3) per-bone \physCOM 
as a proportion of each bone's length, 
and 
(4) whole-body \physCOM as a weighted average of per-bone {\physCOM}s. 
In contrast, our \physCOMp does not require anthropometric measurements and takes into account the full \threeD body shape.}

\begin{figure}[ht]
\centerline{
\includegraphics[width=\linewidth]{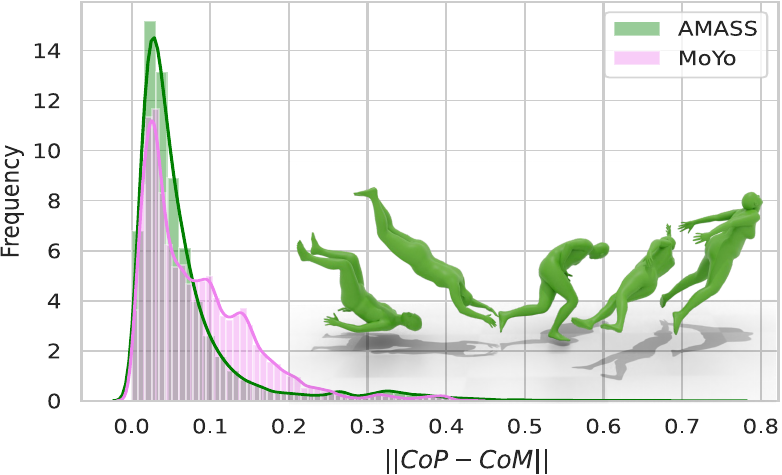}
}
\caption{\newcamready{Distribution of $\mathcal{L}_{\text{stability}}$ in \textcolor{OliveGreen}{AMASS} and \textcolor{VioletRed}{MoYo}. Both peak at $\sim0$, 
motivating using an $L_{2}$ formulation. 
Bottom right: Unstable long-tail poses from \textcolor{OliveGreen}{AMASS}.}}
\label{figure:ipman:moyo_amass_dist}
\end{figure}

%% file: supplementary/sections_Supmat/stability_loss_justification.tex
\subsection{Stability Loss}

\newcamready{The suggested classic definition uses a binary stability criterion, \ie, the \physCOM ``just'' projects either inside or outside the  \physBOS. 
This is discontinuous %
with sparse gradients.} 

\newcamready{Since \physCOP lies inside \physBOS, our L2 loss is a ``soft'' version that approximates the classic definition, but has two key benefits:  
(1) it is continuous and fully differentiable, and, 
(2) it informs about the \emph{degree} of instability.
The distribution of $\mathcal{L}_{\text{stability}}$ in \cref{figure:ipman:moyo_amass_dist} for both {AMASS} and MoYo datasets peak at $\sim0$, motivating using an $L_{2}$ formulation.}

%% file: supplementary/sections_Supmat/m_naive_m_trig.tex
\subsection{Elements of Stability Analysis: Alternative formulations}

\begin{figure*}[ht]
\centering
\includegraphics[width=\linewidth]{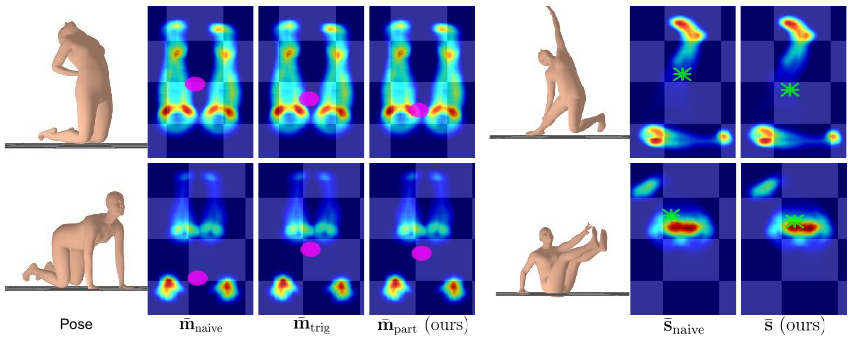}
\vspace*{-0.1in}
\caption{Gravity-projections of different formulations of \physCOM (shown with \textcolor{magenta}{pink}) and \physCOP (shown with \textcolor{green}{green}) are shown with estimated pressure maps. Our proposed \physCOMp captures more accurate body mass distribution because it takes into account part-specific mass contributions. Similarly, our \physCOP leverages the pressure maps rather than binary contact.}
\label{figure:com_cos_ablation}
\end{figure*}

Computation of the ``\physExplainCOM'', \physCOM, must be efficient and differentiable.
The \physCOM could be naively approximated as the mean vertex position of a mesh:
\begin{eqnarray}
    \bar{\mathbf{m}}_{\text{naive}}=\frac{1}{N_V} \sum_{i=1}^{N_V} \bs{v}_i \text{.}
\end{eqnarray}
However, the \smpl and the \smplx body models have a non-uniform vertex distribution across the surface. 
There are a disproportionate number of vertices on the face and hands compared to the body. For instance, roughly half 
of \smplx's vertices lie on the head. Consequently, $\bar{\mathbf{m}}_\text{naive}$ is dominated by face and hand vertices. %
 
\newcamready{A better formulation is the mean of uniformly sampled surface points: 
\begin{eqnarray}
    \bar{\mathbf{m}}^{u}_{\text{naive}}=\frac{1}{N_U} \sum_{i=1}^{N_U} \bs{v}_i \text{.}
\end{eqnarray}
}

Another formulation computes the 
average of the mesh triangle face centroids weighted by the face area: 
\begin{eqnarray}
    \bar{\mathbf{m}}_{\text{trig}} = \frac{\sum_{i=1}^{N_F} A_i \bar{F_i}}{\sum_{i=1}^{N_F} A_i} \text{,}
\end{eqnarray}
where $A_i$ denotes the area and $ \bar{F_i} = \frac{1}{3} (\bs{v}_{i_1}^\top + \bs{v}_{i_2}^\top + \bs{v}_{i_3}^\top)$ the centroid of face $\bs{F}_{i}$. 
\newcamready{The problem with these approaches is that they assume that mass, $M$, is proportional to surface area, $S$, which is a poor approximation.}

\newcamready{Our proposed \physCOMp formulation addresses this by $(1)$ uniformly sampling vertices on the SMPL mesh and $(2)$ taking part-specific mass contributions into account. Our \physCOMp computes mass from volume, $\mathcal{V}$, via the standard density 
equation, 
$M=\rho\mathcal{V}$}. \cref{table:alt_com_comp} compares the \physCOM error across different formulations of \physCOM \wrt ground-truth \physCOM obtained using Vicon PiG. \physCOMp significantly outperforms all baselines. \refFig{figure:com_cos_ablation} shows an intuitive qualitative comparison between all formulations of \physCOM.

\input{supplementary/tables_Supmat/alt_com_comparison}

Similarly, for ``\physExplainCOP'' (\physCOP), a simple heuristic used in previous works detects binary contact by thresholding body vertices using their Euclidean distance from the ground plane. However, such contact lacks information about the pressure distribution and assigns equal weight to all contact vertices. Moreover, binary contact is not differentiable and is therefore generally used at test-time~\cite{Hassan2019prox,Villegas_2021_ICCV,zanfir2018monocular,zhang2020phosa} or for data preprocessing~\cite{Hassan2021posa,Zhang_2021_ICCV}, not during training. In contrast, our \physCOP formulation is fully differentiable and takes the inferred pressure distribution of the body-floor contact into account. As shown in \reffig{figure:com_cos_ablation}, the naive \physCOP suffers from equally weighting all binary-contact whereas our \physCOP better represents the pressure profile of the body-ground contact.  

\subsection{Ablation of ground losses}

\begin{figure}[ht]
\centering
\includegraphics[width=\linewidth]{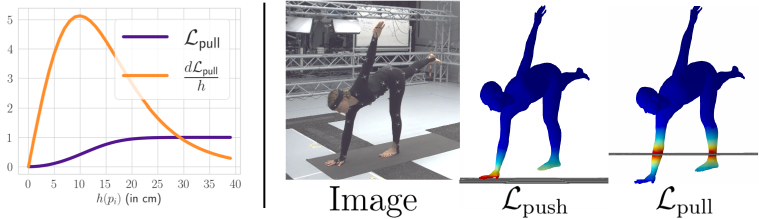}
\caption{\textit{Left}: The gradient of $\mathcal{L}_{\text{pull}}$ decays gradually with $h(p_i)$; vertices with $h(p_i)\geq20$cm contribute minimally to back-propagation. \textit{Right}: Effect of $\mathcal{L}_{\text{push}}$ and $\mathcal{L}_{\text{pull}}$; high $\mathcal{L}_{\text{push}}$ results in inaccurate floor contact, high $\mathcal{L}_{\text{pull}}$ results in penetrations. }
\label{figure:method_losses}
\end{figure}

\newtext{Instead of having a threshold to restrict $\mathcal{L}_{\text{pull}}$ only to vertices close to the ground, we chose a soft version of the loss to ensure full differentiability. However, as shown in \cref{figure:method_losses} (left), the loss gradient decays with height and vertices with $h(v_i) \geq 15$ cm contribute minimally during back-propagation.   
Further, we 
study 
the impact of $\mathcal{L}_{\text{push}}$ and $\mathcal{L}_{\text{pull}}$ in 
\cref{table:rebuttal:ablationPushPull} and \cref{figure:method_losses}-right. 
The terms complement each other and are more effective when used jointly ($\mathcal{L}_{\text{ground}}$).}

\input{supplementary/tables_Supmat/lpull_lpush_ablations}

%% file: supplementary/tables_Supmat/alt_com_comparison.tex
\begin{table}[ht]
    \centering
    \small
    \rowcolors{3}{}{lightgray}
    \renewcommand{\arraystretch}{1.2} %
    \setlength{\tabcolsep}{4pt}
    \begin{adjustbox}{width=\linewidth}
    \begin{tabular}{@{}l|c|c|c|c@{}}
    \Xhline{3\arrayrulewidth}
    
     & $\bar{\mathbf{m}}_{\text{naive}}$ & $\bar{\mathbf{m}}^{u}_{\text{naive}}$ & $\bar{\mathbf{m}}_{\text{trig}}$ & \physCOMp ($\bar{m}$) \\
    \hline
    \physCOM error $\downarrow$ & 264.1 mm & 68.5 mm & 70.0 mm &  \textbf{53.3 mm}\\  %
    \hline
    \Xhline{3\arrayrulewidth}
    \end{tabular}
    \end{adjustbox}
     \caption{
     \newcamready{Comparison of various \physCOM formulations}.
    }
    \label{table:alt_com_comp}
\end{table}

%% file: supplementary/tables_Supmat/lpull_lpush_ablations.tex
\begin{table}[h]
\begin{center}
    \small
    \renewcommand{\arraystretch}{1.0}
    \begin{tabular}{@{}l|ccc @{}}
    \Xhline{3\arrayrulewidth}
         \small{\textbf{Method}} &\small{\textbf{MPJPE}  $\downarrow$} &\small{\textbf{PMPJPE} $\downarrow$}   & \small{\textbf{PVE}   $\downarrow$}   \\
    \hline
  $\text{\hmr}^*$~\cite{Kanazawa2018_hmr} & 82.5 & 48.2 & 92.3 \\ %
  $\text{\hmr}^*$~\cite{Kanazawa2018_hmr}$+\mathcal{L}_{\text{push}}$ & 85.4 & 49.0  & 96.6 \\ 
  $\text{\hmr}^*$~\cite{Kanazawa2018_hmr}$+\mathcal{L}_{\text{pull}}$ & 88.0 & 48.8 & 99.4 \\ 
  $\text{\hmr}^*$~\cite{Kanazawa2018_hmr}$+\mathcal{L}_\text{ground}$ & 80.9 & 47.8 & 89.9  \\
    \Xhline{3\arrayrulewidth}
    \end{tabular}
\caption{Ablation for $\mathcal{L}_{\text{push}}$ and $\mathcal{L}_{\text{pull}}$ on the RICH~\cite{RICH} dataset.}
\label{table:rebuttal:ablationPushPull}
\end{center}
\end{table}

%% file: supplementary/sections_Supmat/implementation_details.tex
\newcamready{We integrate our 
intuitive-physics terms in both 
an optimization-    %
and 
a  regression-based %
method for three reasons:
(1) the community heavily uses both method types, %
(2) our terms generalize and benefit both types, despite their differences, and 
(3) our terms also work with 
different body models;
\smplX (used by \nameMethodO) and \smpl (used by \nameMethodR).}

\subsection{\nameMethod Implementation Details}

\subsubsection{\nameMethodR.}

Similar to previous methods 
\cite{Kocabas_SPEC_2021, kolotouros2019spin, Mueller:CVPR:2021, jiang2020multiperson}, 
we take the widely used \hmr~\cite{Kanazawa2018_hmr} architecture to analyze the effect of adding our proposed \IPterms. 
Note that, while \hmr is not the most recent method, it is widely used as a backbone. As such, it provides a consistent foundation for evaluation and comparison.
Our goal here is to isolate and evaluate the effect of adding intuitive physics.
Such terms should then be readily applicable to other HPS regression frameworks.

The \hmr regressor estimates the camera translation $\mathbf{t}^c$ and \smpl parameters (pose, %
global orientation, and shape) in the camera coordinates assuming 
$\camrot=\bs{I}_3$ and $\bodytransl=\mathbf{0}$. We initialize the \hmr model using pretrained weights provided by SPIN~\cite{kolotouros2019spin} and finetune both \nameMethodR and $\hmr$ on the same datasets; namely \rich~\cite{RICH}, Human3.6M~\cite{ionescupapavaetal2014}, MPI-INF-3DHP~\cite{Mehta2017Monocular3H}, COCO~\cite{Lin2014MicrosoftCC}, MPII~\cite{andriluka14cvpr} and LSP~\cite{Johnson10, Johnson11}. In the main paper, we call the baseline as $\text{\hmr}^*$ which uses the same training datasets and hyperparameters as \nameMethodR, albeit with the exception of the proposed \IPterms. We follow the same training schedule, data augmentation and hyperparameters as SPIN~\cite{kolotouros2019spin} but do not use in-the-loop optimization. We use the Adam optimizer with learning rate of $5e^{-5}$ and finetuning takes 3 epochs ($\sim 8$ hours) on a Nvidia Tesla V100 GPU. 

We set the hyperparameters $\alpha=100$, $\gamma=10$ for the per-vertex pressure $\rho_i$, $\alpha_1=1.0$, $\alpha_2=0.15$ for the $\mathcal{L}_\text{pull}$ term and $\beta_1=10.0$, $\beta_2=0.15$ for the $\mathcal{L}_\text{push}$ term. The loss weights are empirically determined to be $\lambda_s=0.01$ and $\lambda_g=0.01$. We borrow the same configuration as \cite{kolotouros2019spin} for all remaining loss weights, namely $\mathcal{\lambda}_\text{2D}$, $\mathcal{\lambda}_\text{3D}$ and $\mathcal{\lambda}_\text{SMPL}$.

\newtext{
\rich~\cite{RICH} contains sequences with an uneven ground-plane. For training \nameMethodR, we therefore sample a subset of the \rich dataset where subjects mainly interact with an even ground plane (see \cref{table:rich_sample}). In the Train/Val sequences, we use camera 0 for validation and cameras 1-5 for training. }

\input{supplementary/tables_Supmat/rich_seq_list}

\subsubsection{\nameMethodO.}
For \nameMethodO, we extend the baseline optimization-based method \smplifyXMC~\cite{Mueller:CVPR:2021}. We use the same configuration as \smplifyXMC and only add extra hyperparameters for the proposed \IPterms. Both methods are initialized with the same presented pose from the \nameYOGIData dataset. We extract 2D keypoints from images using MediaPipe~\cite{Lugaresi2019MediaPipeAF}.   

Same as \nameMethodR, we set the hyperparameters $\alpha=70$, $\gamma=10$ for the per-vertex pressure $\rho_i$, $\alpha_1=1.0$, $\alpha_2=0.15$ for the $\mathcal{L}_\text{pull}$ term and $\beta_1=10.0$, $\beta_2=0.15$ for $\mathcal{L}_\text{push}$ term. The loss weights are empirically determined to be $\lambda_s=10000$ and $\lambda_g=10000$.

%% file: supplementary/tables_Supmat/rich_seq_list.tex
\begin{table}[ht]
    \centering
    \tiny
    \setlength{\tabcolsep}{4pt}
    \begin{adjustbox}{width=0.8\linewidth}
    \begin{tabular}{@{}c|c@{}}
    \Xhline{3\arrayrulewidth}
    Train/Val & Test \\
    \hline
    'Pavallion\_000\_yoga2' & `Pavallion\_002\_yoga1' \\
    'Pavallion\_000\_yoga1' & `Pavallion\_013\_yoga2' \\
    'Pavallion\_006\_yoga1' & `ParkingLot2\_014\_pushup2' \\
    'Pavallion\_018\_yoga1' & `ParkingLot1\_005\_pushup1' \\

    \Xhline{3\arrayrulewidth}
    \end{tabular}
    \end{adjustbox}
     \caption{Training, validation and test sequences in the \rich dataset containing an even ground. 
    }
    \label{table:rich_sample}
\end{table}

%% file: supplementary/sections_Supmat/bose_calculation.tex
\subsection{Evaluation Metrics}
\subsubsection{\physBOS Error (\physBOSE) calculation.}

Recall that the ``\physExplainBOS'' (\physBOS) is defined by the convex hull of the contact regions.
Since computing this can be computationally inefficient, we reformulate the \physBOSE computation to test if projection of the \physCOM, $g(\bar{\mathbf{m}}_{\text{part}})$, on the ground plane can be represented as a convex combination of the gravity-projected contact vertices $C$. To this end,  we solve the linear equation system via standard linear programming using interior point methods~\cite{Andersen2000TheMI}:
\begin{eqnarray}
    &\min_{\mathbf{a}} &\|\mathbf{a}^\top C - \bar{\mathbf{m}}_{\text{part}}\| \\
    &\text{s.t.} &a_i \in \mathbf{a} \geq 0 \\
    & &\sum a_i =1
\end{eqnarray}
where $\mathbf{a}^\top C = a_1 \bs{c}_1 + \dots + a_n \bs{c}_n$ for the points $\bs{c}_i$ in $C$. If the system has a solution, $g(\bar{\mathbf{m}}) \in \mathcal{C}(C)$ holds, otherwise $g(\bar{\mathbf{m}})$ is not in the convex hull of $C$, \ie $g(\bar{\mathbf{m}}) \notin \mathcal{C}(C)$.

\begin{figure}[ht]
\centerline{
\includegraphics[width=\linewidth]{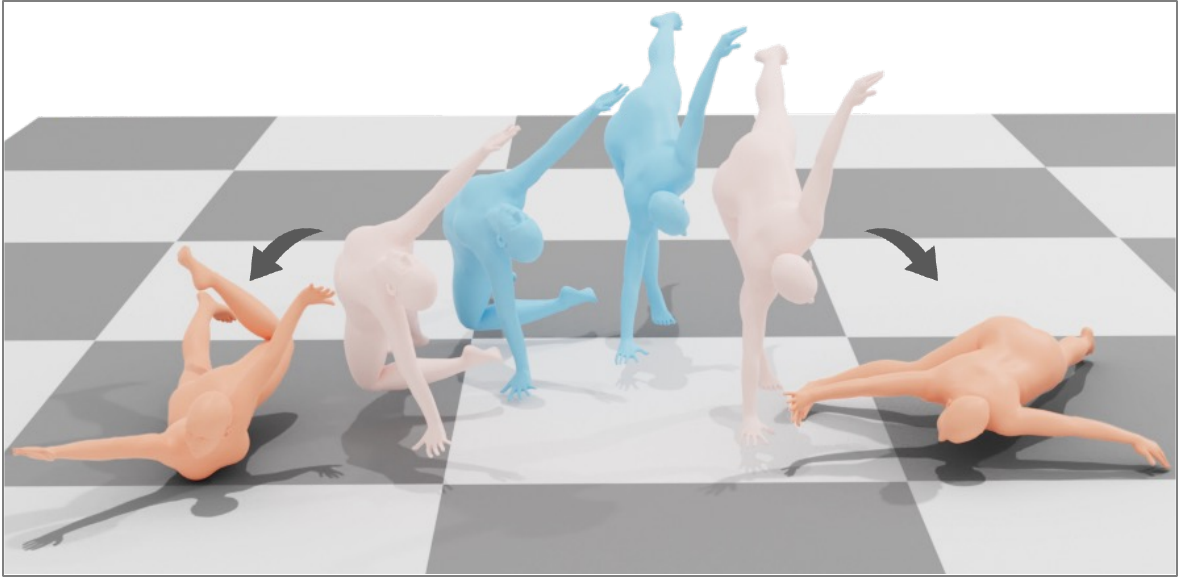}
}
\caption{\newcamready{Stability evaluation using the ``Bullet'' physics engine. Meshes produced by the baseline method~\cite{Mueller:CVPR:2021} (in \textcolor{orange}{orange}) topple but \nameMethodO's meshes (in \textcolor{cyan}{cyan}) remain stable after physics simulation.}}
\label{figure:ipman:physics_sim}
\end{figure}

%% file: supplementary/sections_Supmat/extra_qualitative_results.tex
\subsection{Qualitative Results}

\begin{figure*}[ht]
\centerline{
\includegraphics[width=\linewidth]{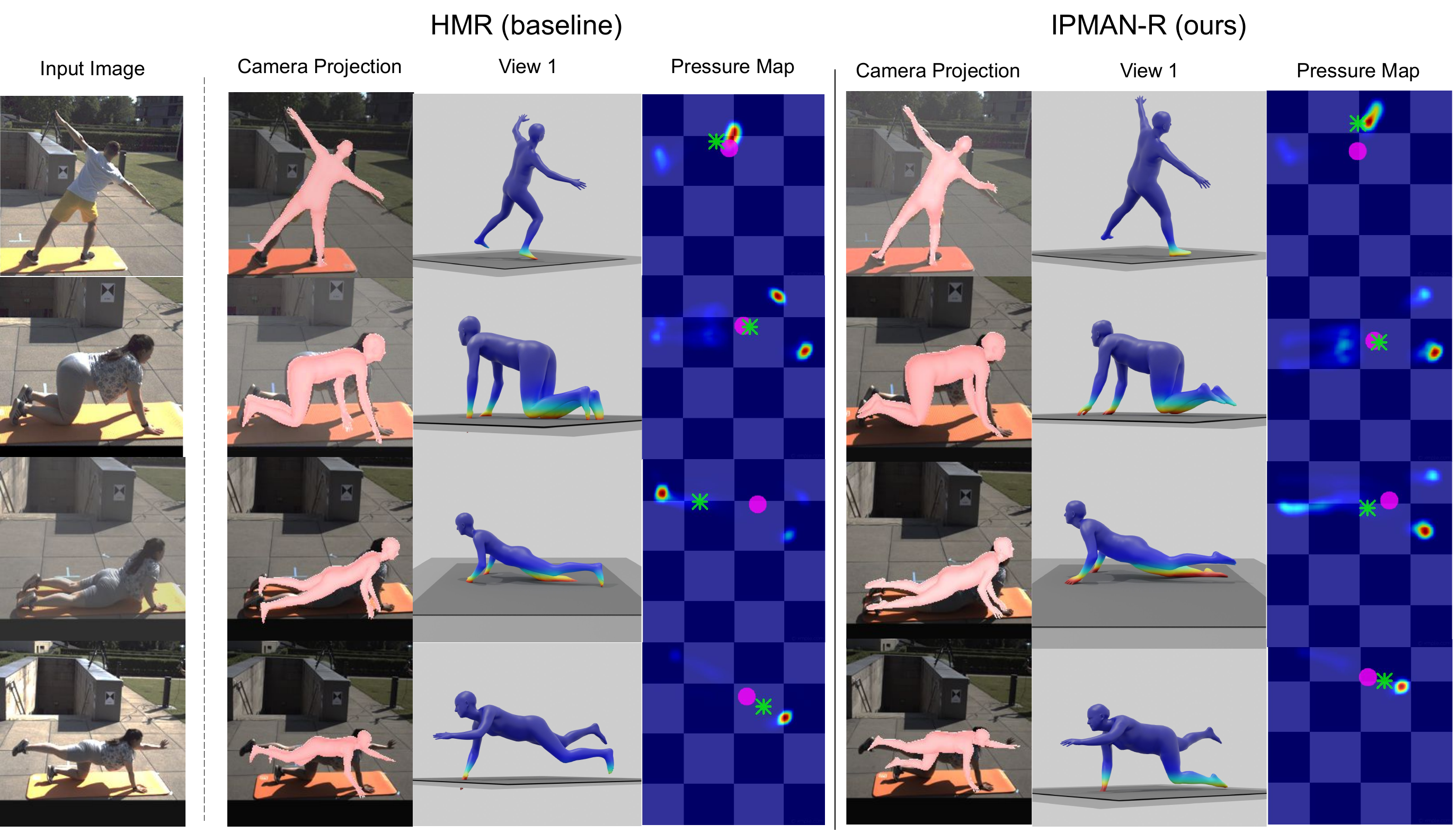}
}
\caption{Additional qualitative evaluation of \nameMethodR on \rich. The first column shows the input images of a subject doing various sports poses. The second and third block of columns show the results of \hmr (baseline) and \nameMethodR, respectively. In each block, the first image shows the estimated mesh 
overlaid
on the image. The next three images show different views of the estimated mesh in the world frame. The \textcolor{green}{green} sphere illustrates the \physCOM.}
\label{figure:ipmanr_qual_result_supmat}
\end{figure*}

\begin{figure*}[ht]
\centerline{
\includegraphics[width=\linewidth]{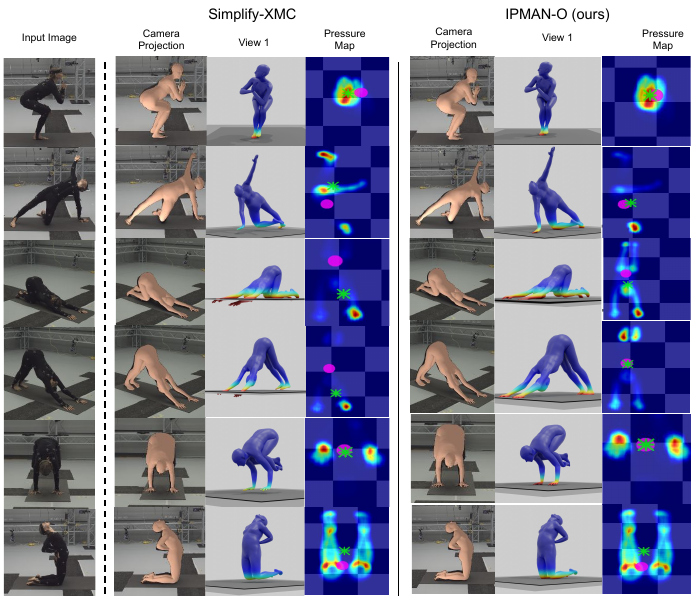}
}
\caption{Qualitative evaluation of \nameMethodO on \yogiIOI. In the first column, the input images of a subject doing yoga poses. The second and third blocks show the results of the \smplifyXMC and \nameMethodO respectively. In each block, the first and second column show the estimated mesh projected into the image and from a second view. The last images show the pressure map with the \physCOM (in \textcolor{magenta}{pink}) and the \physCOP (in \textcolor{green}{green}).
}
\label{figure:ipmano_qual_result_supmat}
\end{figure*}

\refFig{figure:ipmanr_qual_result_supmat,figure:ipmano_qual_result_supmat} show supplemental qualitative results for \nameMethodR and \nameMethodO, respectively. 

%% file: supplementary/sections_Supmat/physics_simulation.tex
\newcamready{
Current physics engines 
are incompatible 
with \HPS methods, 
as they approximate \smpl bodies with rigid convex hulls and are non-differentiable. 
However, using them for posthoc stability evaluation of the estimated meshes is possible. Specifically, we evaluate \nameMethodO and \mbox{SMPLify-XMC}~\cite{Mueller:CVPR:2021} by first, using V-HACD convex decomposition~\cite{mamou2016volumetric} of the estimated body meshes and then by simulating physics as in \cite{hasson2019obman, Tzionas2016CapturingHI} via the ``Bullet'' physics engine~\cite{pyBullet}. We measure the displacement of the human mesh after 100 physics simulation steps; a small displacement denotes a stable pose and vice versa. \nameMethodO produces $14.8\%$ 
more 
stable bodies than the baseline~\cite{Mueller:CVPR:2021}; see \cref{figure:ipman:physics_sim}.}  

%% file: supplementary/sections_Supmat/biomechanical_eval.tex
\newtext{We use the pressure field defined in Eqn.~2 of the main paper to compute per-point pressure on the \smpl mesh. With this, the pressure heatmap is estimated by summing the per-point pressure projected to the ground-plane. Note that we recover relative pressure as we do not assume availability of ground-truth body mass or anthropometric measurements.} 

\newtext{
To measure the overlap of the inferred pressure heatmap \wrt the ground-truth, we compute the intersection-over-union (IOU) between the two. However, the ZEBRIS pressure sensor captures pressure measurements in the range 10-1200 KPa. Depending upon the contact area and the weight of the subject, some poses may fall outside this range. For instance, a person lying-down only exerts 1-5 kPa of pressure on the ground. To account for this, we tune the sensitivity of our pressure field for every pose and report mean of the best per-sample IOU.}

\newtext{
We measure accuracy of our \physCOP by simply computing the Euclidean distance \wrt ground-truth. We call this as \physCOP error. Again, we report mean of the best \physCOP error after tuning the sensitivity of our inferred pressure field.} 

\newtext{
The \physCOM error is similar to the \physCOP error, albiet in \threeD. It measures the Euclidean distance between the estimated and ground-truth \physCOM recovered from Vicon Plug-in Gait. \Cref{table:biomech_eval} presents summary results showing that our inferred pressure, \physCOP and \physCOM agrees with the ground-truth.}

\input{supplementary/tables_Supmat/biomech_eval}

%% file: supplementary/tables_Supmat/biomech_eval.tex
\begin{table}[ht]
    \centering
    \small
    \rowcolors{3}{}{lightgray}
    \renewcommand{\arraystretch}{1.2} %
    \setlength{\tabcolsep}{4pt}
    \begin{adjustbox}{width=\linewidth}
    \begin{tabular}{@{}l|l c|c@{}}
    \Xhline{3\arrayrulewidth}
    \multirow{2}{*}{} & \multicolumn{2}{c|}{\textbf{Pressure}} & \textbf{\physCOM} \\

     & \textbf{mIOU} & \textbf{\physCOP error (mm)} & \textbf{\physCOM error (mm)} \\
    \hline
    \nameMethod (Ours) & 0.32 & 57.3 & 53.3 \\  %
    \hline
    \Xhline{3\arrayrulewidth}
    \end{tabular}
    \end{adjustbox}
     \caption{
     Quantitative evaluations of our estimated pressure, \physCOP and \physCOM \wrt ground-truth in \nameYOGIData.
    }
    \label{table:biomech_eval}
\end{table}

%% file: supplementary/sections_Supmat/04_smplifyx_IPMAN_O.tex
\section{\nameMethodOnew (\NEWWW{Extension of} \smplifyX).}

To further explore the effect of our intuitive-physics terms, we extend the optimization method \smplifyX~\cite{Pavlakos2019_smplifyx} \NEWWW{and name this \nameMethodOnew (note that this is different from the main paper's \nameMethodO that extends \smplifyXMC)}. We fit the \smplx body model to \twoD image keypoints starting from mean pose and shape while exploiting the ground-truth ground plane. Adapted from \smplifyX \cite{Pavlakos2019_smplifyx}, we minimize the objective
\begin{align}
E(\bs{\beta}, \bs{\theta}, \bs{\psi}, \camtransl) =
& E_{J2D} + 
E_{\theta} + \lambda_{\beta} E_{\beta} + \lambda_{\psi} E_{\psi} + 
\nonumber 
\\
&
\lambda_{\alpha} E_{\alpha} + \lambda_{\mathcal{C}} E_{\mathcal{C}} + 
\\
& 
\lambda_{s} E_{\text{stability}} + \lambda_{g} E_{\text{ground}}
\nonumber 
\text{.}
\end{align}
The energy term $E_{J2D}$ denotes the \twoD re-projection error whereas the remaining terms $E_{\theta} = \lambda_{\theta_b} E_{\theta_b} + \lambda_{\theta_f} E_{\theta_f} + \lambda_{\theta_h} E_{\theta_h}$ represent various priors for body, face, and hand pose. $E_{\beta}$, $E_{\psi}$, $E_{\alpha}$ and $E_{\mathcal{C}}$ are prior terms for body shape, expression, extreme bending and self-penetration (see \cite{Pavlakos2019_smplifyx} for details).
 $E_{S}$ and $E_G$ are the stability and ground contact losses. The results in \reftab{table:sota_optimization_x} show a clear improvement.

 Note that \smplifyX estimates the body's global orientation $\bodyori$ and the camera translation $\camtransl$, while camera rotation $\camrot$ and body translation $\bodytransl$ remain zero. In order to apply our \IPterms, we use the ground-truth camera rotation $\mathbf{R}^c_w$ and translation $\mathbf{t}^c_w$ to transform the estimated mesh from camera to world coordinates. We empirically find that applying the \IPterms to the final stage of optimization in \smplifyX gives more accurate results than applying them to all stages. We hypothesize that this could be due to having a better body initialization before applying the \IPterms.

\input{supplementary/tables_Supmat/TAB_00_IPMAN_O_X}

%% file: supplementary/tables_Supmat/TAB_00_IPMAN_O_X.tex
\begin{table}[t]
    \centering
    \small
    \rowcolors{3}{}{lightgray}
    \renewcommand{\arraystretch}{1.2} %
    \setlength{\tabcolsep}{4pt}
    \begin{tabular}{@{}l|l c c @{}}
    \Xhline{3\arrayrulewidth}
   
    \multirow{2}{*}{\textbf{Method}}  & \multicolumn{3}{c}{\textbf{\rich \cite{RICH}}}  \\
    
     & \textbf{MPJPE $\downarrow$} & \textbf{PVE $\downarrow$} & \textbf{\physBOSE (\%) $\uparrow$} \\
    \hline
    \smplifyX~\cite{Pavlakos2019_smplifyx}  & 268.6 & 228.3 & 96.9   \\
    \nameMethodOnew (Ours) & \textbf{240.9} & \textbf{217.1} & \textbf{98.0} \\  %
    \hline
    \Xhline{3\arrayrulewidth}
    \end{tabular}
    \caption{
            \nameMethodOnew compared to the optimization method of \cite{Pavlakos2019_smplifyx} on \rich~\cite{RICH}.
    }
    \label{table:sota_optimization_x}
\end{table}

%% file: supplementary/sections_Supmat/3dpw_eval.tex
\section{Evaluation on \threedpw}

\input{supplementary/tables_Supmat/TAB_01_3DPW}

\threedpw~\cite{vonmarcard_eccv_2018_3dpw} is an outdoor dataset containing pseudo ground-truth \smpl and camera parameters recovered using IMU sensors attached to the actors. As also noted in \cite{yuan2021glamr}, we find that the ground plane in \threedpw is inconsistent. In fact, two subjects in the same scene can be supported by different ground-planes in the world coordinates. Additionally, \threedpw primarily contains dynamic poses like walking, climbing stairs, parkour, etc. Due to these reasons, \threedpw does not satisfy the core assumptions of \nameMethod.
Nevertheless, we report results on \threedpw to show that the \IPterms do not degrade performance for such datasets; in fact, we see a slight improvement in performance
as illustrated in  \refTab{table:ipman_r_3dpw}. %
This makes \nameMethod applicable to everyday motion without needing special care.

%% file: supplementary/tables_Supmat/TAB_01_3DPW.tex
\begin{table}[ht]
    \centering
    \small
    \rowcolors{3}{}{lightgray}
    \renewcommand{\arraystretch}{1.2} %
    \setlength{\tabcolsep}{4pt}
    \begin{tabular}{@{}l|l c @{}}
    \Xhline{3\arrayrulewidth}
   
    \multirow{2}{*}{\textbf{Method}}  & \multicolumn{2}{c}{\textbf{\threedpw~\cite{vonmarcard_eccv_2018_3dpw}}}  \\
    
     & \textbf{MPJPE $\downarrow$} & \textbf{PMPJPE $\downarrow$} \\
    \hline
    SPIN~\cite{kolotouros2019spin}  & 97.2 & 59.6   \\
    \nameMethodR (Ours) & \textbf{96.8} & \textbf{57.1} \\  %
    \hline
    \Xhline{3\arrayrulewidth}
    \end{tabular}
     \caption{
            \nameMethodR compared to the regression method of \cite{kolotouros2019spin} on \threedpw~\cite{vonmarcard_eccv_2018_3dpw}.
    }
    \label{table:ipman_r_3dpw}
\end{table}